\documentclass[acmtog, authorversion, nonacm]{acmart}

\AtBeginDocument{%
  }

\usepackage{graphicx}
\usepackage{booktabs}
\usepackage{mathtools}
\usepackage{amsfonts}
\usepackage[inline]{enumitem}
\usepackage{algorithm}
\usepackage{algpseudocode}
\usepackage{nicefrac}
\usepackage{overpic}
\usepackage{tikz}
\usepackage{wrapfig}
\usepackage{float}
\usepackage{cleveref}

\def\eg{\emph{e.g.}}
\def\ie{\emph{i.e.}}

\usepackage{xcolor}

\definecolor{myred3}{RGB}{250,3,3} 
\definecolor{myred2}{RGB}{180,3,3} 
\definecolor{myred1}{RGB}{110,3,3} 
\definecolor{BlueMATLAB}{HTML}{0072BD}
\definecolor{GreenMATLAB}{HTML}{77AC30}

\begin{document}

\title{GSEdit: Efficient Text-Guided Editing of 3D Objects via Gaussian Splatting}

\author{Francesco Palandra}
\authornote{Both authors contributed equally to this research.}
\author{Andrea Sanchietti}
\authornotemark[1]
\affiliation{%
  \institution{Sapienza University of Rome}
  \city{Rome}
  \country{Italy}
}

\author{Daniele Baieri}
\email{baieri@di.uniroma1.it}
\orcid{0000-0002-0704-5960}
\affiliation{%
  \institution{Sapienza University of Rome}
  \city{Rome}
  \country{Italy}}

\author{Emanuele Rodol\`a}
\email{rodola@di.uniroma1.it}
\orcid{0000-0003-0091-7241}
\affiliation{%
  \institution{Sapienza University of Rome}
  \city{Rome}
  \country{Italy}}

\renewcommand{\shortauthors}{Trovato et al.}

\begin{abstract}
We present \textbf{GS-Edit}, a pipeline for text-guided 3D object editing based on \textbf{G}aussian \textbf{S}platting models. 
Our method enables the editing of the style and appearance of 3D objects without altering their main details, all in a matter of minutes on consumer hardware. 
We tackle the problem by leveraging Gaussian splatting to represent 3D scenes, and we optimize the model while progressively varying the image supervision by means of a pretrained image-based diffusion model. The input object may be given as a 3D triangular mesh, or directly provided as Gaussians from a generative model such as DreamGaussian.
GS-Edit ensures consistency across different viewpoints, maintaining the integrity of the original object's information. 
Compared to previously proposed methods relying on NeRF-like MLP models, GS-Edit stands out for its efficiency, making 3D editing tasks much faster. 
Our editing process is refined via the application of the SDS loss, ensuring that our edits are both precise and accurate. 
Our comprehensive evaluation demonstrates that GS-Edit effectively alters object shape and appearance following the given textual instructions while preserving their coherence and detail.


\end{abstract}


\begin{teaserfigure}
    \includegraphics[width=\columnwidth]{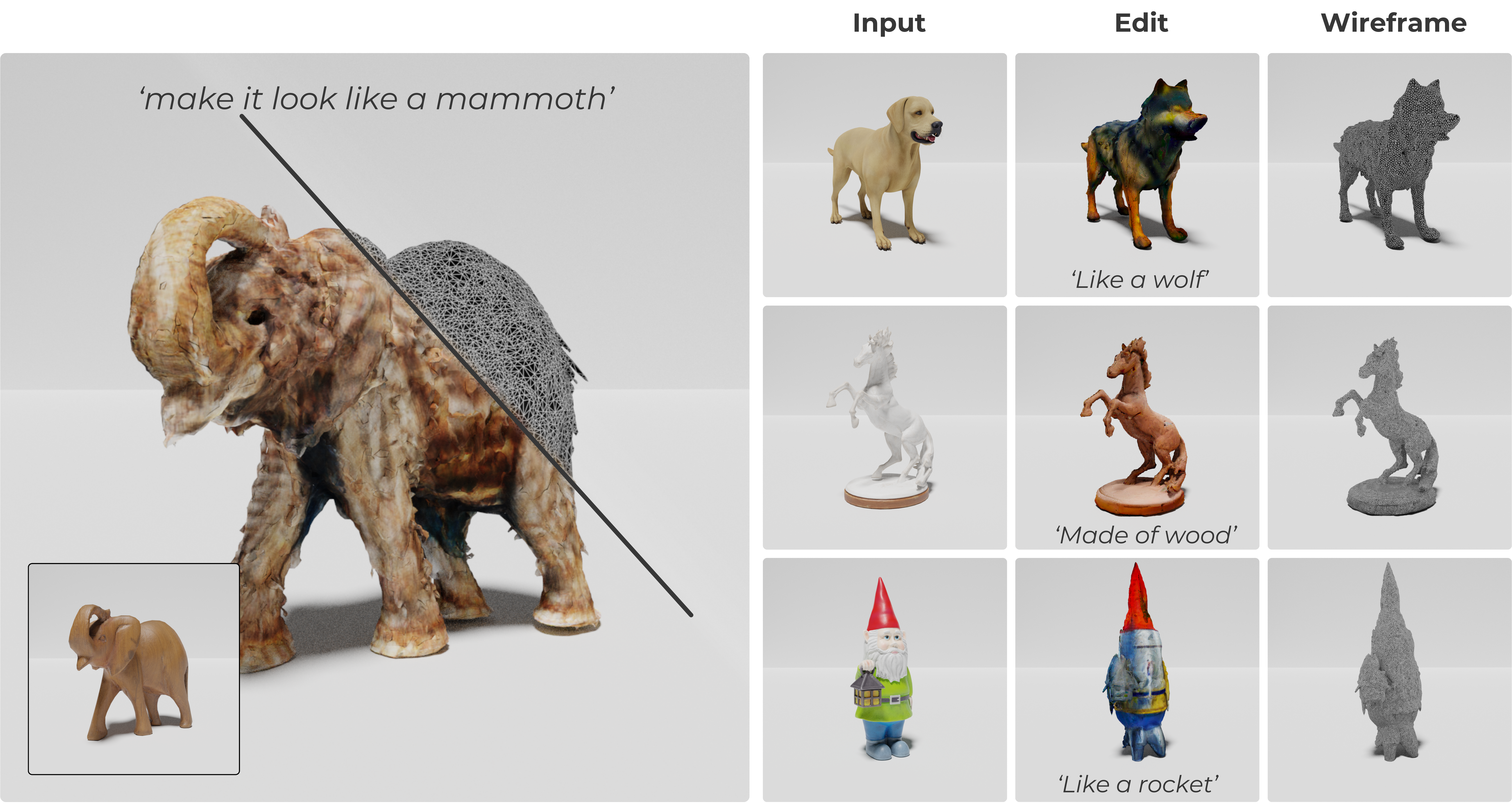}
    \caption{Results of 3D object editing with GSEdit. This method can take as input either meshes or point clouds, and perform an edit in a few minutes, guided by a textual prompt and a diffusion model.}
    \label{fig:teaser}
\end{teaserfigure}


\maketitle

\section{Introduction and related work}


\begin{figure*}[t!]
    \centering
    \includegraphics[width=\linewidth]{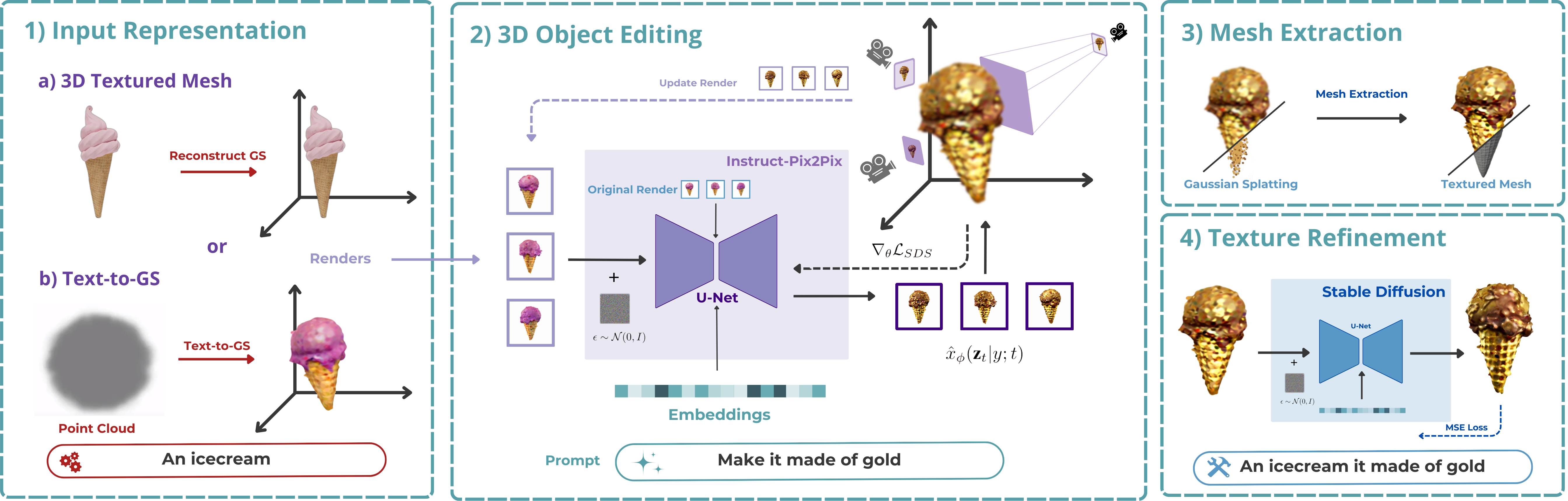}
    \caption{The GSEdit pipeline. 1) It starts by taking a mesh a) or a pretrained Gaussian splatting model b) as input. During this stage, 20 different views of the shape are rendered, and if the input is a mesh, the renders are used to encode it in a GS model. 2) The editing phase consists of picking a camera, rendering the scene from that camera, running a step of Instruct-Pix2Pix \cite{brooks2023instructpix2pix} to apply the edit, and optimizing the SDS \cite{poole2022dreamfusion} loss. Once the editing is complete, the mesh extraction 3) and texture refinement 4) steps introduced in DreamGaussian~\cite{tang2023dreamgaussian} are carried out.}
    \label{fig:pipeline}
\end{figure*}

The fast-paced progress of research in the fields of deep learning and artificial intelligence in recent years brought significant advancement in 3D vision and graphics. \citet{mildenhall2020nerf} were the first to propose the idea of optimizing volumetric color/density functions encoded by deep neural networks (referred to as NeRF, for Neural Radiance Fields) via differentiable volumetric rendering. When the output renders from multiple views are optimized to match ground truth views of some environment, the model changes accordingly to encode the geometry and appearance of the scene. 

Due to the exceptional impact it had on the field, NeRF inspired a variety of contributions improving it in several aspects. One of these is rendering time: encoding the color and density functions in a deep and wide MLP, which required millions of queries to render a single view, led to sub-optimal inference performance. The simplest way to approach this problem is to bake the neural functions into efficient discrete structures such as voxel grids \cite{Garbin2021FastNeRFHN, hedman2021snerg} or octrees \cite{yu2021plenoctrees, fourierplenoct}. Alternatively, one could change the rendering model, making the volume integral more efficient \cite{autoint, wu2021diver}, or using smaller MLPs in combination with space decomposition \cite{Reiser2021KiloNeRFSU} or multi-resolution hash encoding \cite{mueller2022instant}. Recently,~\citet{kerbl20233d} proposed to represent scenes in a hybrid fashion with Gaussian splatting (GS in the following), which employs a finite set of 3D Gaussians to define the continuous color and density fields, while achieving impressive efficiency with an ad-hoc differentiable rasterization algorithm. 

Besides a completely new way of approaching the inverse rendering problem, NeRF provided a novel, general framework to handle geometry in an implicit fashion, thus bypassing all issues derived from explicit structures such as triangle meshes. For instance, automated 3D editing/generation tasks were always limited by the complexity of handling changes in the discrete topology, while NeRF-like models allowed to perform these tasks in the neural network's input/weight space. Two main categories of approaches can be identified: a) using conditional latent codes, either from pretrained image GAN/VAE priors \cite{GRAF, nerfvae, piGAN2021, EG3D, cai2022pix2nerf, gu2022stylenerf} or jointly trained \cite{editnerf, Wang2021CLIPNeRFTD}, and b) editing/generating the ground truth multi-view images with pretrained diffusion models. This idea was first introduced in DreamFusion by~\citet{poole2022dreamfusion}, and it rapidly became the state of the art in this line of research due to the popularity and performance of diffusion models. Subsequent work experimented with combining this framework with Instant-NGP~\cite{lin2023magic3d,melaskyriazi2023realfusion} as well as training volumetric fields to output latent features to be fed into a Stable Diffusion decoder~\cite{latentnerf, stablediff}. DreamGaussian \cite{tang2023dreamgaussian} was the first model for text-to-3D using GS models.


Our work focuses on exploiting the efficiency of GS models in order to define an efficient pipeline for text-guided 3D object editing. Achieving accurate and efficient text-guided editing of 3D objects would have notable implications: mainly, it would mean greater empowerment for 3D artists, allowing them to perform complex, fine grained modifications to their digital environments quickly and automatically. Clearly, this benefit would also similarly reflect on industrial 3D graphics pipelines. Following previous work by~\citet{instructnerf2023}, we rely on a pretrained instance of the Instruct-Pix2Pix (IP2P) diffusion model \cite{brooks2023instructpix2pix} to edit ground truth views given a textual prompt, which we use to fine-tune a GS scene encoding the input object. A similar contribution was proposed by~\citet{voxe}, however, their voxel-based pipeline requires around one hour to run to completion. GS models were previously applied to 3D scene editing by~\citet{GaussianEditor} and~\citet{chen2023gaussianeditor}, with a different focus than our work. Our contributions may be summarized as follows:
\begin{itemize}
    \item We adapt the formulation of the Score Distillation Sampling (\textbf{SDS}) loss~\cite{poole2022dreamfusion} for GS generation proposed by~\citet{tang2023dreamgaussian}, to GS editing, deriving analytical gradients.
    \item We introduce a pipeline for 3D object editing that can perform significant modifications of any input shape, in a handful of minutes, on consumer-grade hardware.
\end{itemize}

\section{Background}

\paragraph{Gaussian Splatting}

The idea of representing 3D scenes as Gaussian functions for reconstruction tasks is due to~\citet{kerbl20233d}. This approach models features (\eg, color) as spatial distribution and employs a fast, differentiable rasterization algorithm to optimize the scene with respect to given ground truth views. 
Unlike traditional NeRF which requires dense sampling and can be computationally expensive, Gaussian Splatting encodes the scene's continuous representation via a finite set of Gaussian functions $\{\Theta_i\}_{i=1}^{n}$, leading to faster training and rendering times while keeping reconstruction quality similar to NeRF-based methods. 
In the initial stages, the scene is approximated via an initial guess $\Theta^0$ based on 2D projection, which is easy to compute for 3D Gaussians. Each Gaussian within this framework is defined by a covariance matrix, $\Sigma_i$ centered around $\mu_i$, embodying the spatial distribution as:
\begin{equation}
    \Theta_i(x) = e^{-\frac{1}{2}(x)^T\Sigma_i^{-1}(x)}
\end{equation}
To introduce variations in visibility and blending across the scene, an opacity value, $\alpha_i$, modulates each Gaussian's influence. In order for the covariance matrix to have physical meaning, we require it to be positive semi-definite, so the authors opted for the (equivalently expressive) following representation, analogous to the configuration of an ellipsoid:
\begin{equation}
    \Sigma_i = R_iS_iS_i^TR_i^T
\end{equation}
Where the rotation and scaling matrices are encoded as a quaternion $r_i \in \mathbb{R}^4$ and a scaling factor $s_i \in \mathbb{R}^3$. Lastly, each Gaussian is associated with a color value $c_i \in \mathbb{R}^3$ to encode appearance.
In order to accurately render color properties, the method refines spherical harmonics (SH) coefficients for color representation within each Gaussian, $\Theta_i$. The iterative nature of Gaussian Splatting allows for dynamic adaptation, including Gaussian modification, creation, and destruction, leveraging stochastic gradient descent for optimization. The opacity values are computed by a sigmoid function, ensuring they remain within a practical range, and the aggregate loss function is composed of an L1 norm and a structural dissimilarity index (D-SSIM):
\begin{equation}
    \mathcal{L} = (1-\lambda) \mathcal{L}_1 + \lambda \mathcal{L}_{D-SSIM}
\end{equation}
Periodically, a densification process prunes Gaussians with opacities below a defined threshold, $\epsilon_{\alpha}$. Notably, the model supports nuanced adjustments such as Gaussian cloning in under-represented regions and subdivision in areas of high detail variance, with scaling adjustments determined by a factor $\phi$. A massive advantage of Gaussian Splatting is its rendering efficiency, replacing volumetric rendering with a tile-based rasterization approach, characterized by 16x16 tiles which not only accelerates computations but also preserves differentiability, enabling effective backpropagation even with extensive Gaussian blending.

\paragraph{Instruct-Pix2Pix}
The success of diffusion models in recent years led to a proliferation of text-based image editing/generation methodologies, which span a broad spectrum from fine-tuning practices, as proposed by~\citet{kawar2023imagic}, to resource-intensive supervised frameworks such as Instruct-Pix2Pix~\cite{brooks2023instructpix2pix}.
The latter is a diffusion model for image editing based on text conditioning. At its core, the authors created a paired training dataset from the combination of a large language model~\cite{brown22language} and a text-to-image model such as Stable Diffusion~\cite{stablediff}. This data is used to train a conditional Diffusion model that outputs an edited image from a source image and a user text prompt. IP2P operates on the latent space since studies have demonstrated that efficiency and quality are improved in Latent Diffusion Models~\cite{https://doi.org/10.48550/arxiv.2204.11824}. Given an input image $x$, a noise $\epsilon \sim \mathcal{N}(0,1)$, scaled according to the timestamp $t$, is added to the latent representation of the input $z = E(x)$ to produce a noisy latent $z_t$. From here a U-Net $\epsilon_\phi$ is trained to predict the noise of $z_t$ given the input prompt $c_T$ and the conditioned image $c_I$ with the aim of minimizing the loss:
\begin{equation}
L=\mathbb{E}_{E(x), E\left(c_I\right), c_T, \epsilon \sim \mathcal{N}(0,1), t}\left[\left\lVert \epsilon-\epsilon_\phi\left(z_t, t, E\left(c_I\right), c_T\right) \right\rVert_2^2\right]
\end{equation}
The model employs a U-Net architecture~\cite{ronneberger2015unet}, renowned for its efficacy in image-to-image translation tasks, to modulate edits as instructed by text embeddings. These embeddings, derived from user-provided textual instructions, guide the model to perform contextually relevant edits while maintaining coherence with the source image. The interaction between the image condition and text embeddings within the U-Net framework enables precise, instruction-aligned modifications, highlighting the model's ability to intuitively adapt to diverse editing tasks.

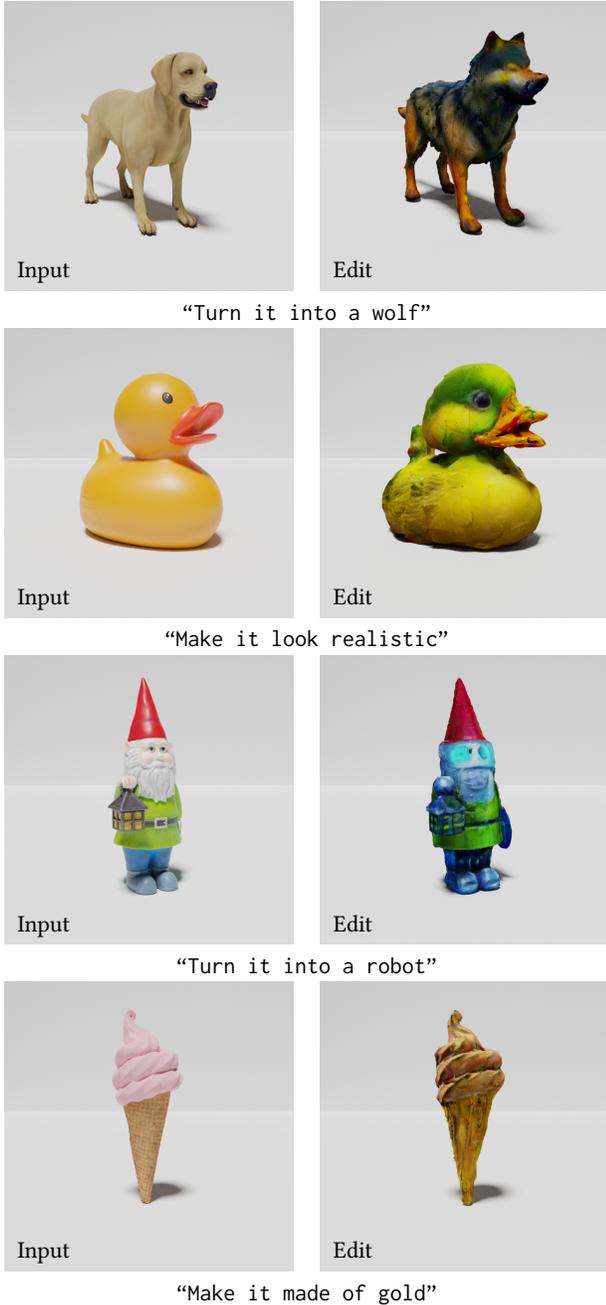
\begin{figure}[h]
    \centering
    \begin{tabular}{c c}
        \begin{picture}(.45\linewidth, .45\linewidth)
            \put(0,0){\includegraphics[width=.45\linewidth]{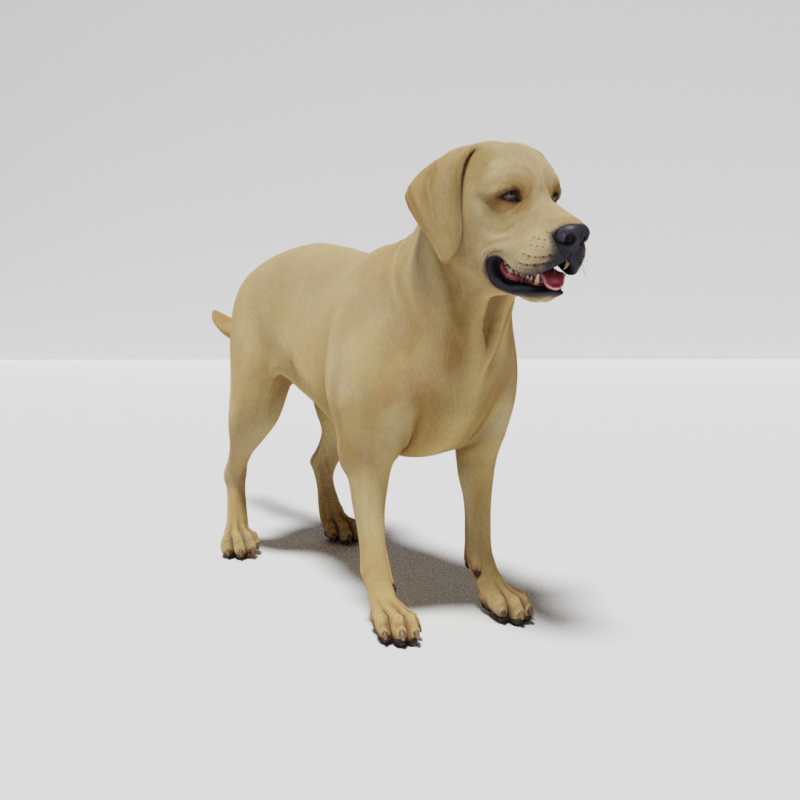}}
            \put(5, 5){Input}
        \end{picture}
        &
        \begin{picture}(.45\linewidth, .45\linewidth)
            \put(0,0){\includegraphics[width=.45\linewidth]{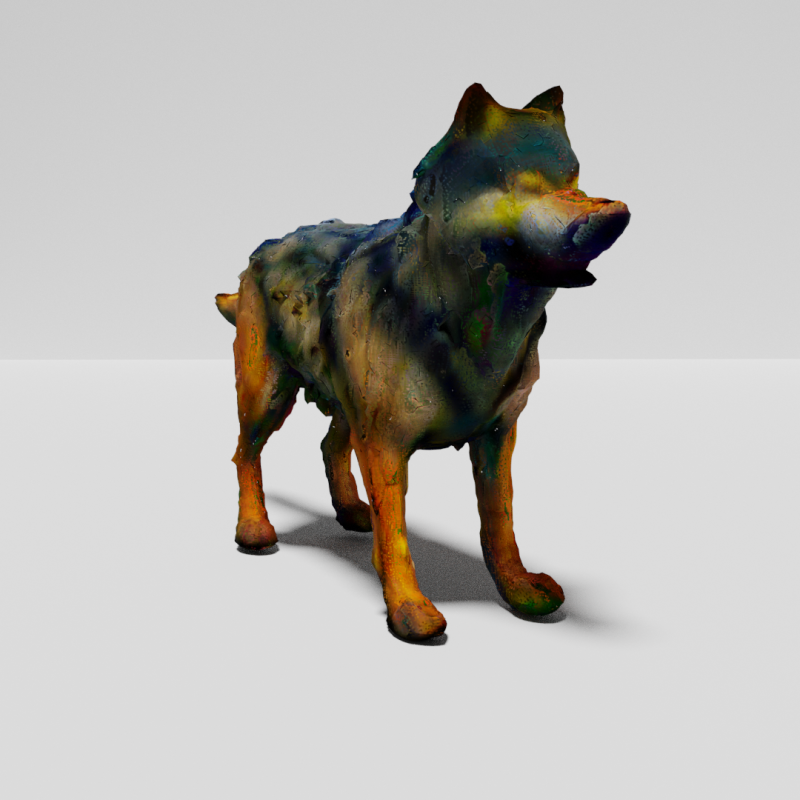}}
            \put(5, 5){Edit}
        \end{picture}
        \\
        \multicolumn{2}{c}{\texttt{``Turn it into a wolf''}} \\
        \begin{picture}(.45\linewidth, .45\linewidth)
            \put(0,0){\includegraphics[width=.45\linewidth]{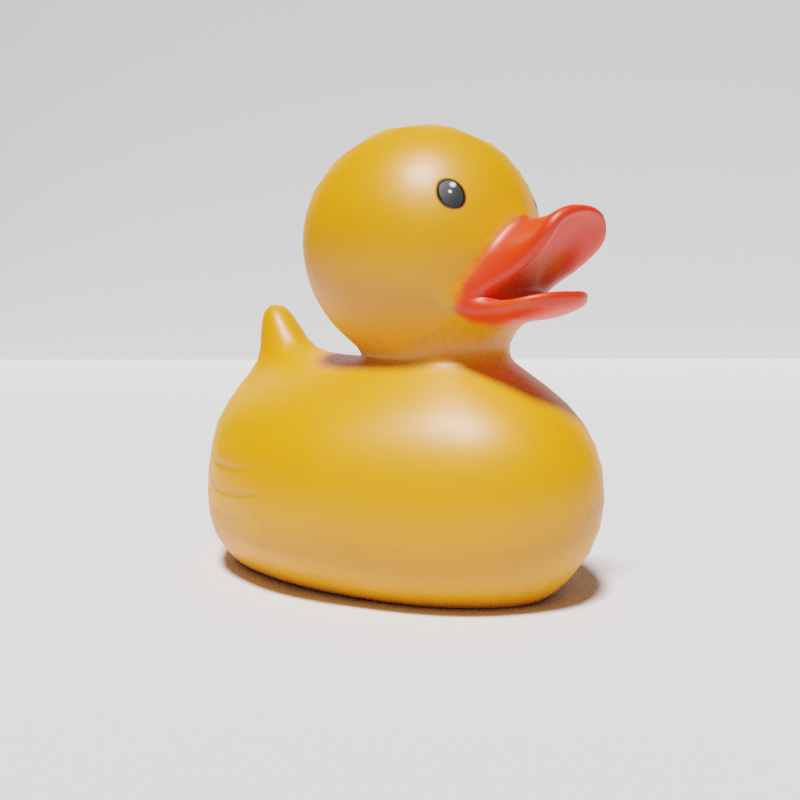}}
            \put(5, 5){Input}
        \end{picture}
        &
        \begin{picture}(.45\linewidth, .45\linewidth)
            \put(0,0){\includegraphics[width=.45\linewidth]{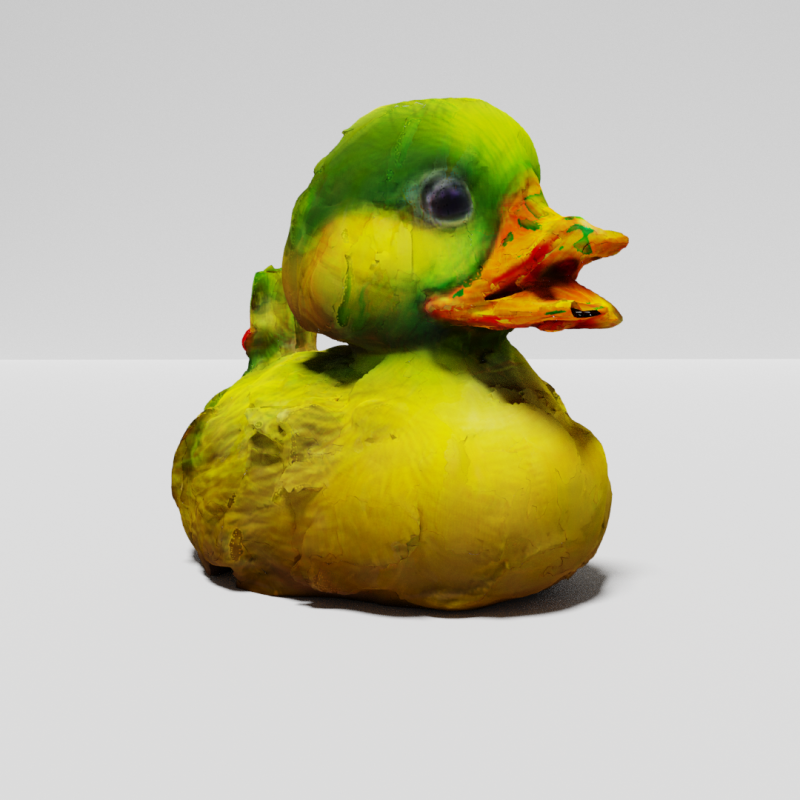}}
            \put(5, 5){Edit}
        \end{picture}
        \\
        \multicolumn{2}{c}{\texttt{``Make it look realistic''}} \\
        \begin{picture}(.45\linewidth, .45\linewidth)
            \put(0,0){\includegraphics[width=.45\linewidth]{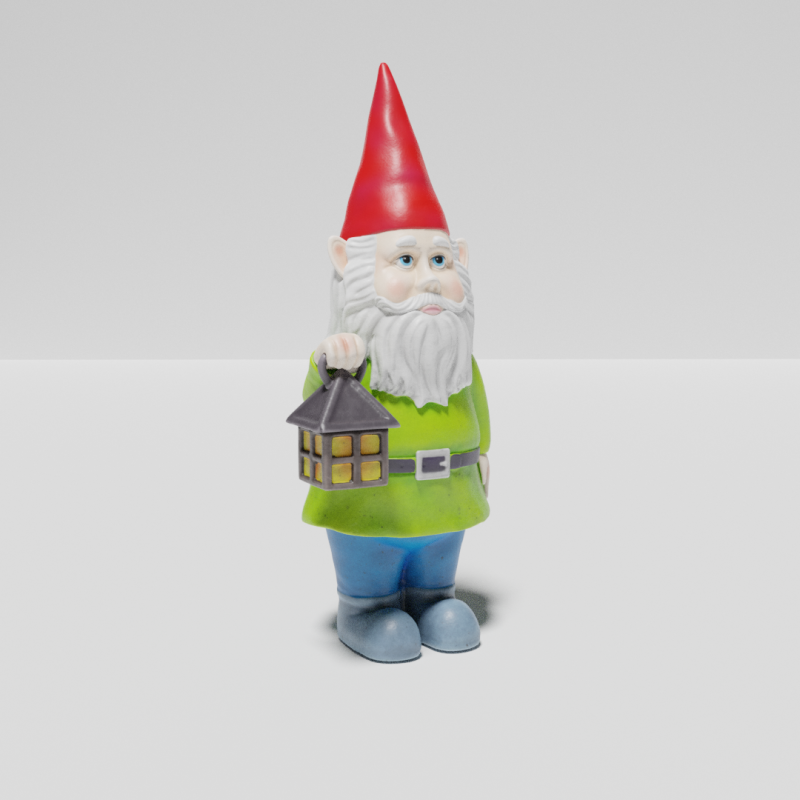}}
            \put(5, 5){Input}
        \end{picture}
        &
        \begin{picture}(.45\linewidth, .45\linewidth)
            \put(0,0){\includegraphics[width=.45\linewidth]{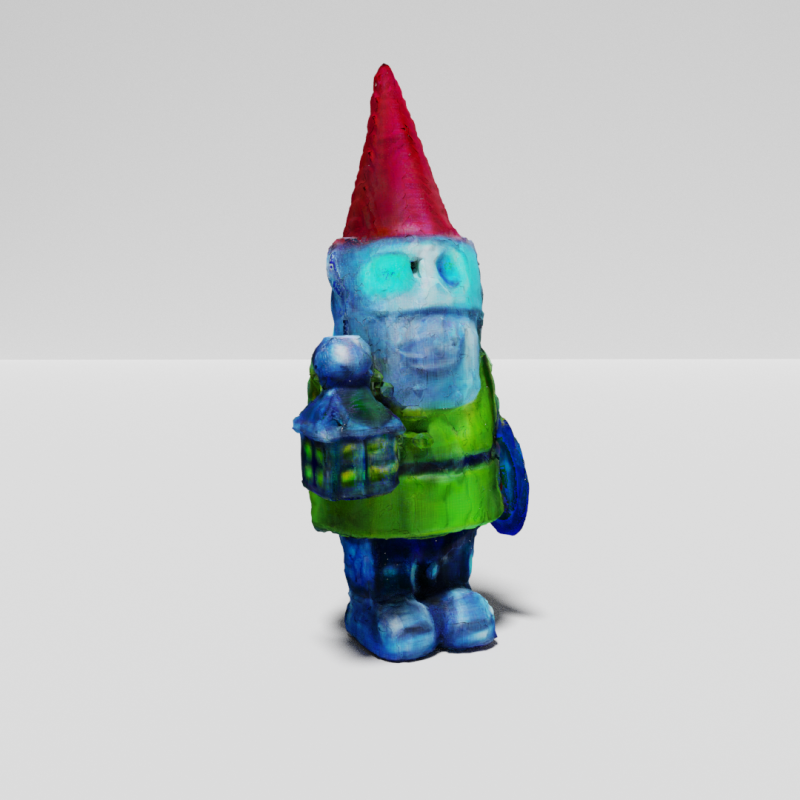}}
            \put(5, 5){Edit}
        \end{picture}
        \\ 
        \multicolumn{2}{c}{\texttt{``Turn it into a robot''}} \\
        \begin{picture}(.45\linewidth, .45\linewidth)
            \put(0,0){\includegraphics[width=.45\linewidth]{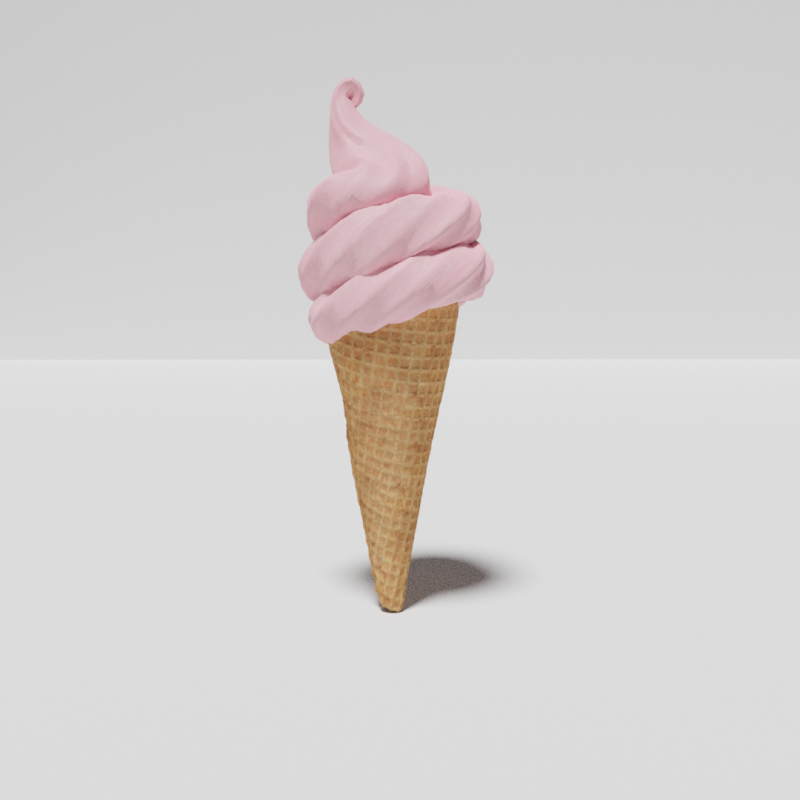}}
            \put(5, 5){Input}
        \end{picture}
        &
        \begin{picture}(.45\linewidth, .45\linewidth)
            \put(0,0){\includegraphics[width=.45\linewidth]{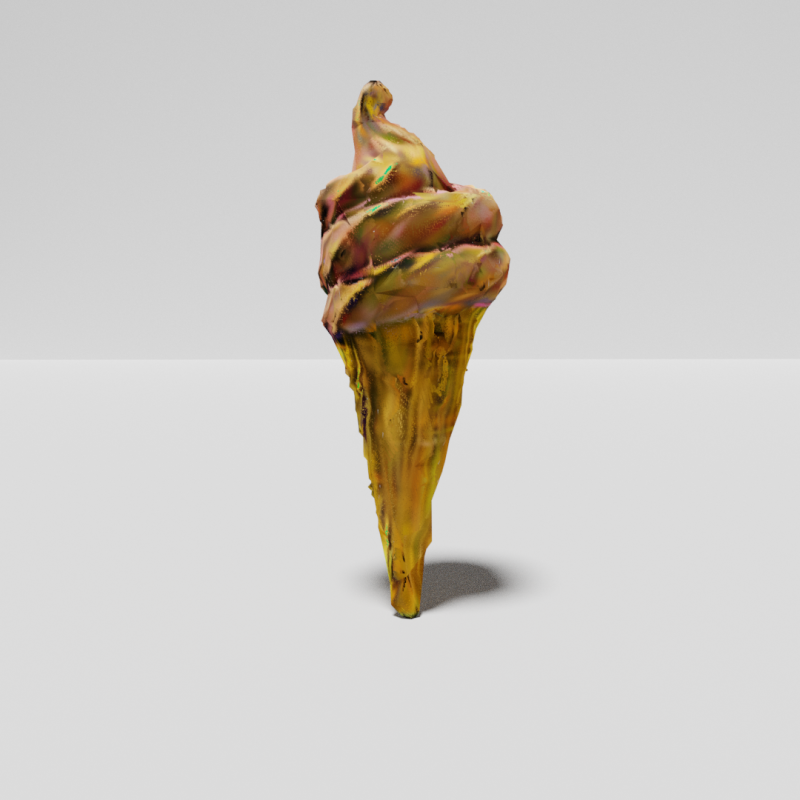}}
            \put(5, 5){Edit}
        \end{picture}
        \\
        \multicolumn{2}{c}{\texttt{``Make it made of gold''}} \\
    \end{tabular}
    \caption{A collection of input/output pairs for our GSEdit pipeline. The results show that our method is flexible enough to handle various settings similarly well, such as changes of object identity (dog $\rightarrow$ wolf, gnome $\rightarrow$ robot) and visual style (realistic duck, golden ice-cream). GSEdit aims to perform the editing while preserving the input's features, such as pose, overall shape, or the hat and lantern of the gnome.}
    \label{fig:collection}
\end{figure}

\section{Method}

Our pipeline runs in four steps, which we summarize in \Cref{fig:pipeline}. In this section, we detail each of them individually.

\subsection{Input Representation}

In order to keep our method as general as possible, we implemented it to support any Gaussian splatting scene as input. However, for the scope of this paper, we are interested in working with single foreground objects; therefore, we consider the following cases as valid inputs for our pipeline: a) GS reconstructions, computed on-the-fly from multi-view renders of a given textured 3D mesh. Foreground/background segmentation in this ``synthetic'' case is trivial, and Gaussian splatting optimization always manages to reconstruct the foreground object alone, without background ``ghostly artifacts''~\cite{Nerfbusters2023} which were common for previous NeRF-like models. b) GS scenes output by generative models such as DreamGaussian~\cite{tang2023dreamgaussian}. We chose DreamGaussian as a generation procedure for part of our experiment scenes due to this model being tailored for 3D object generation, rather than complete scenes. Given a textual prompt, DreamGaussian rapidly generates 3D Gaussians matching the instructions. We provide examples of such inputs for our pipeline in \Cref{fig:pipeline}.

\subsection{3D Object Editing}
For the editing part, we use the same cameras placed for the generation of the Gaussian Splatting to retrieve in each training step the current render of the scene\footnote{If the input is generated by DreamGaussian, we place random cameras and generate multi-view renders accordingly, as we do in the case of training a GS from scratch.}. At each step, we sample a random camera $p$ orbiting the object center and render the RGB image $I^p_{RGB}$. This render is subsequently processed through the IP2P model along with the input textual prompt for editing. We adopt a singular step adjustment strategy akin to the one employed in Dream Gaussian~\cite{tang2023dreamgaussian}, focusing on incremental and directional enhancements within the CLIP space to preserve the integrity of the scene. Once all of the rendered images are processed, the parameters of the Gaussians are updated via differentiable rendering to match the current ground truth renders. Afterward, images to feed into the IP2P model are updated by rendering the new GS scene. This cyclical process of capturing, editing, and updating methodically evolves the original scene, allowing a gradual and controlled transition until the editing state has converged (\ie, IP2P does not apply significant modifications to the current state). 

\begin{figure*}[t]
    \centering
    \begin{tabular}{c c c c c}
        \begin{picture}(.18\linewidth, .18\linewidth)
            \put(0,0){\includegraphics[width=.18\linewidth]{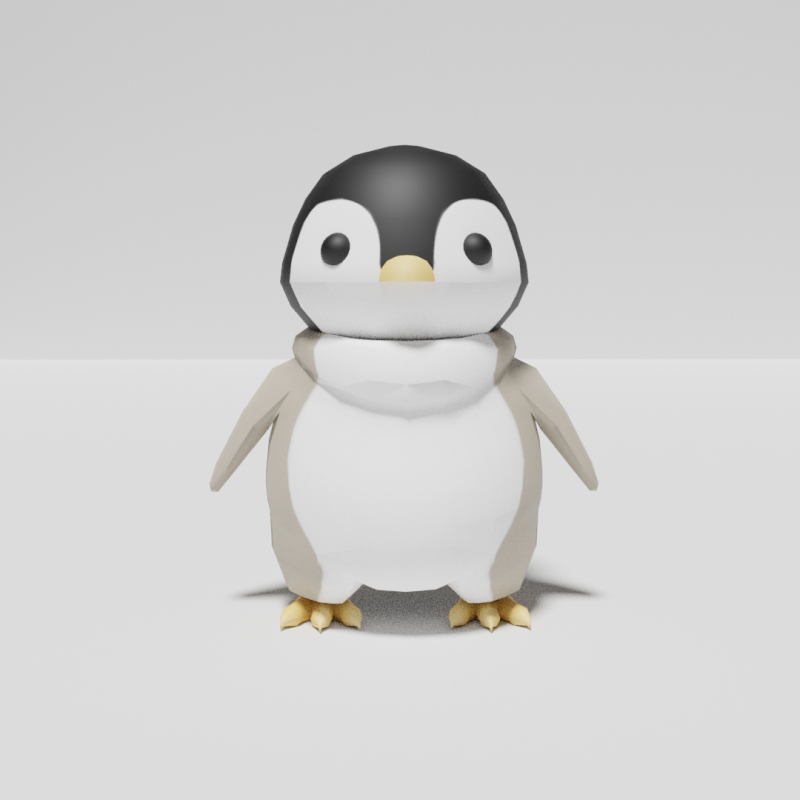}}
            \put(5, 5){Input}
        \end{picture}
        &
        \begin{picture}(.18\linewidth, .18\linewidth)
            \put(0,0){\includegraphics[width=.18\linewidth]{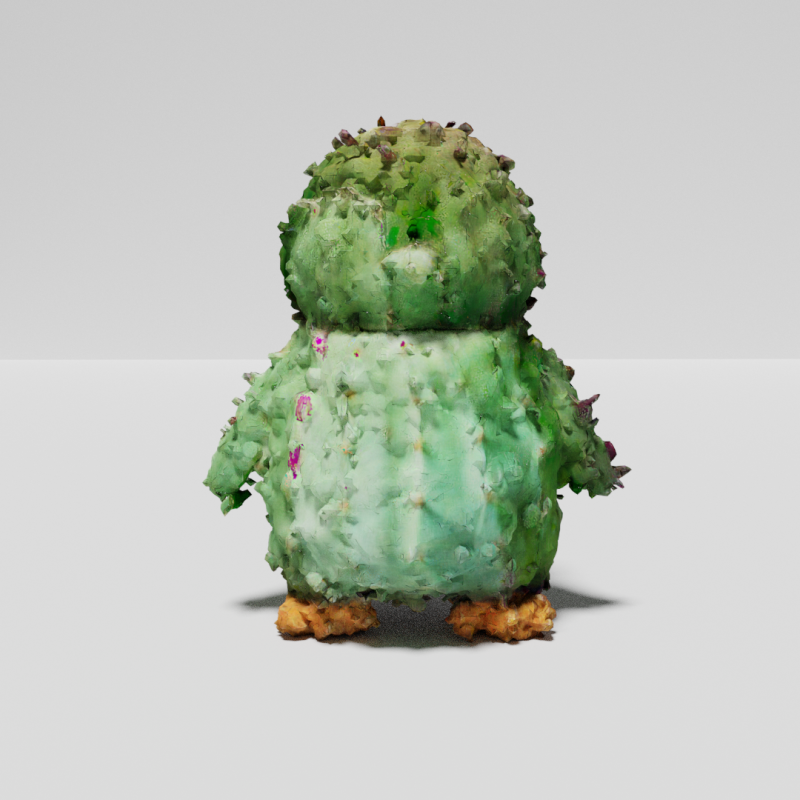}}
            \put(5, 5){Into a cactus}
        \end{picture}
        &
        \begin{picture}(.18\linewidth, .18\linewidth)
            \put(0,0){\includegraphics[width=.18\linewidth]{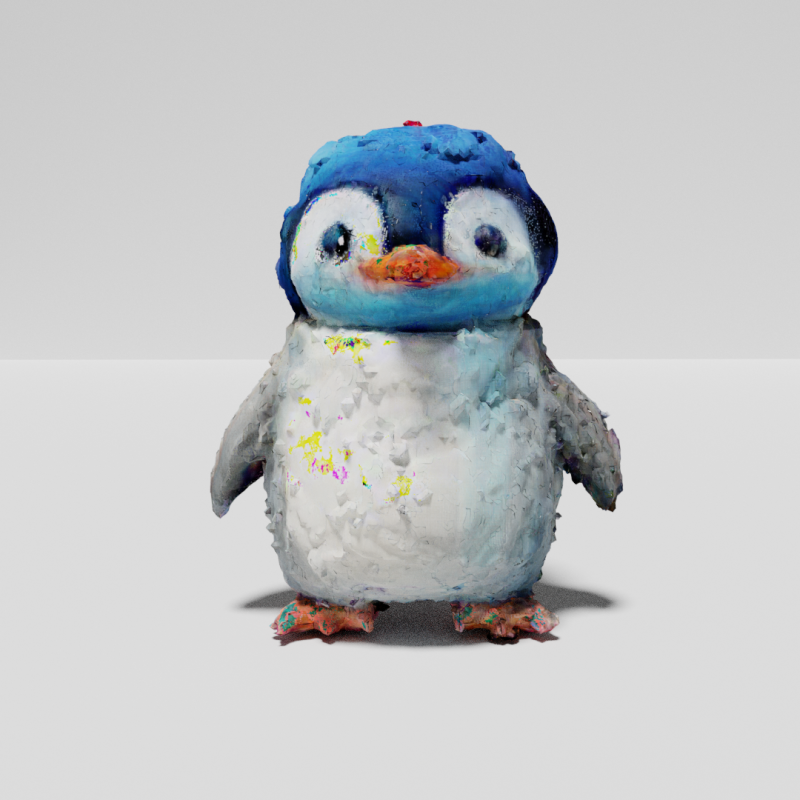}}
            \put(5, 5){Put a hat}
        \end{picture}
        &
        \begin{picture}(.18\linewidth, .18\linewidth)
            \put(0,0){\includegraphics[width=.18\linewidth]{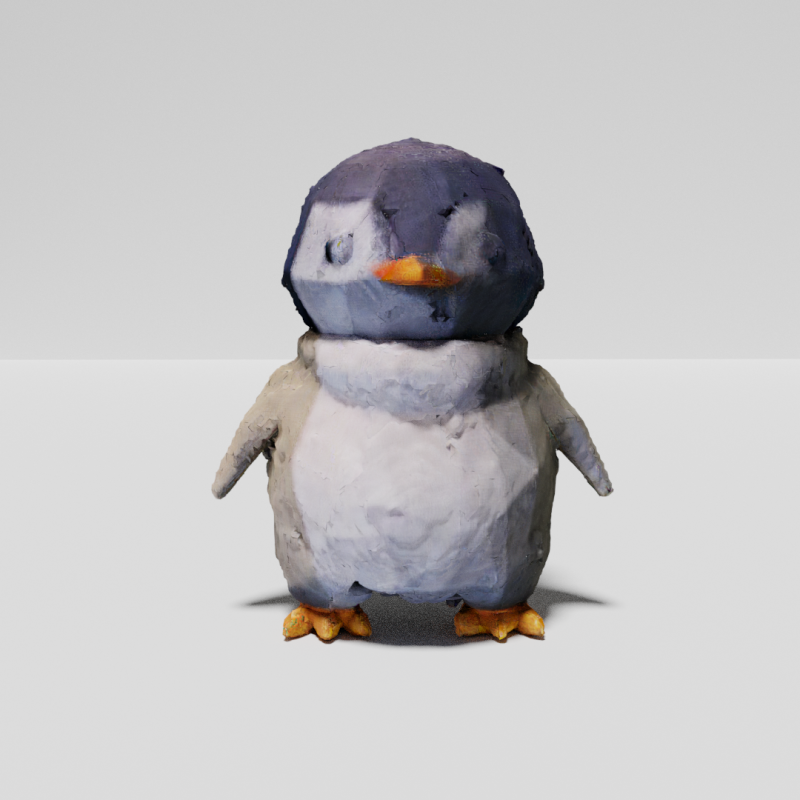}}
            \put(5, 5){Low poly}
        \end{picture}
        &
        \begin{picture}(.18\linewidth, .18\linewidth)
            \put(0,0){\includegraphics[width=.18\linewidth]{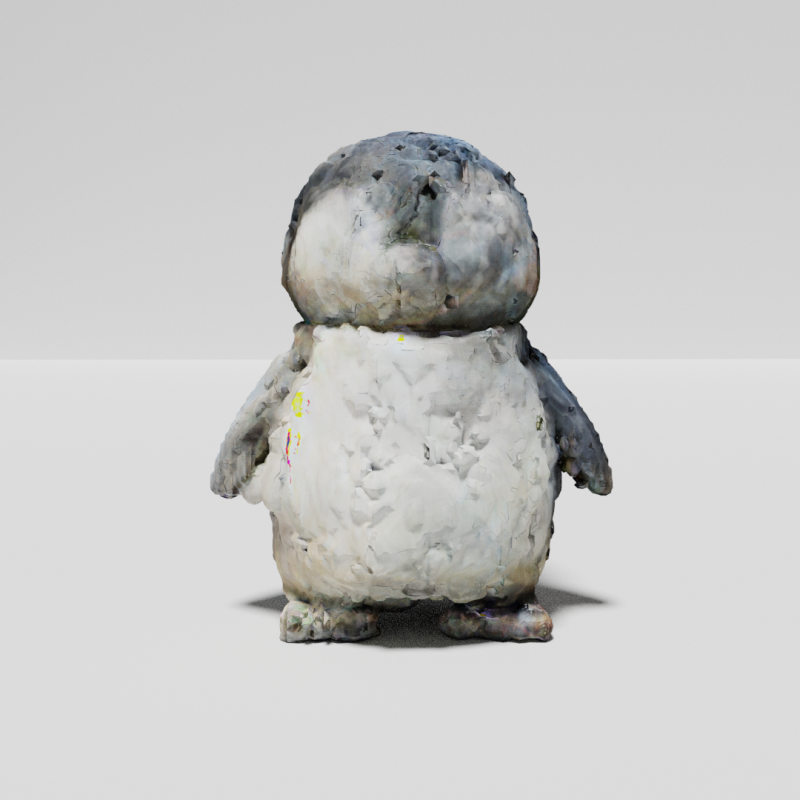}}
            \put(5, 5){Marble}
        \end{picture}
    \end{tabular}
    \caption{Examples of applications of GSEdit with fixed input mesh, varying the editing instruction. The results show the ability of our model to change object identity while keeping original features (cactus), adding simple features (hat), editing ``artistic'' style (lowpoly), and material (marble).}
    \label{fig:penguin}
\end{figure*}
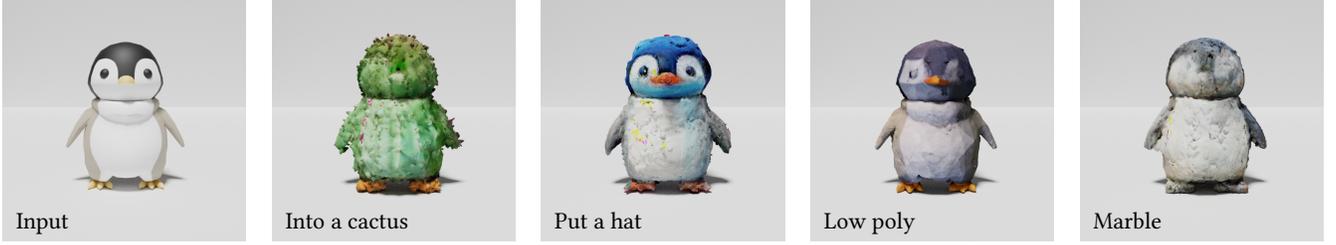

\begin{figure}[h]
    \centering
    \includegraphics[width=.98\linewidth]{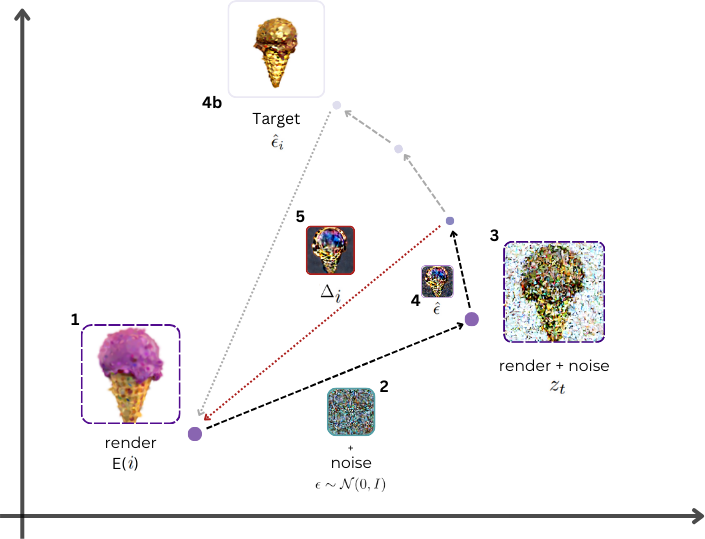}
    \caption{A visualization of how the edits performed by the U-Net of IP2P are exploited to guide the edit.}
    \label{fig:gradient-transform}
\end{figure}

\paragraph{Gradient Backpropagation}
We now explain how the Score Distillation Loss~\cite{poole2022dreamfusion} (SDS) can be adapted to GS editing. \Cref{fig:gradient-transform} summarizes this process: given a timestep $t$, a noise $\epsilon \sim  \mathcal{N}(0,I)$ (2) is added to the $i$-th rendered image encoded with IP2P (1), resulting in a noisy image $z_t$ (3). Afterward, the pretrained IP2P U-Net $\epsilon_{\phi}$ tries to predict the amount of noise $\hat{\epsilon}$ (4) in the input image $z_t$ given the original image $c_I$ and the text embeddings for the editing instruction $c_T$:
\begin{equation}
    \hat{\epsilon} = \epsilon_{\phi} (z_t ; t, c_I, c_T)
\end{equation}

The text embedding has the purpose of guiding the reconstruction toward an image that is similar to the original one but with elements of the text. The noise residual (5) is finally computed as the difference between the predicted noise $\hat{\epsilon}$ and the initial noise $\epsilon$:
\begin{equation}
    \Delta_i = \hat{\epsilon} - \epsilon
\end{equation}
The residual contains all the edits made by the U-Net to match the input $z_{t}$ and the prompt $c_{T}$, and it depends on the output of our GS model, so it can be used to backpropagate the error through the rasterizer. The final formula that encapsulates all the operations applied during a training step is the following:
\begin{equation}
    \nabla_{\Theta} \mathcal{L}_{\text{SDS}} = \mathbb{E}_{t \sim \mathcal{U}(1, T)}\left[w(t)\Delta_i \dfrac{\partial z_t}{\partial \Theta}\right]
\end{equation}
where $w(t)$ is a weighting function that depends on the timestep $t$. We clip $t$ in a $[t_{min}, t_{max}]$ range and we linearly decrease this value as proposed by~\citet{huang2023dreamtime} so that the amount of noise added to the latent render $z_i$ is greater in the initial steps and decreases over time.

\subsection{Mesh Extraction and Texture Refinement}
To efficiently extract and refine a polygonal mesh from the edited point cloud, we leverage the method introduced in \cite{tang2023dreamgaussian}. Their method is divided into a first part that extracts the mesh, and a second part that refines textures. The former consists of two steps: 
\begin{enumerate}
    \item \textbf{Local Density Query.} To extract the mesh geometry, they rely on Marching Cubes \cite{marching_cubes}. The first step is to reduce the number of Gaussians by dividing the space into $16^{3}$ and removing all those Gaussians whose center is located outside each local block. Then, they query a $8^3$ dense grid inside each block and sum up the weighted opacity of each of the $k$ Gaussian in the block:
    \begin{equation}
        d(x) = \sum_{i=1}^{k}\alpha_{i}\exp\left(-\frac{1}{2}(x - x_{i})^{T}\Sigma_{i}^{-1}(x-x_{i})\right)
    \end{equation}
    where $\Sigma_{i}$ is the covariance matrix derived from scaling $s_{i}$ and rotation $q_{i}$. To extract the final mesh, a threshold is applied to extract the mesh surface through Marching Cubes. Finally, they post-process the mesh with a decimation and remeshing step.
    \item \textbf{Color Back-projection}. This step consists of unwrapping the mesh UV coordinates and then assigning colors based on the renders of the object. Each pixel from the RGB render can be back-projected to the texture image based on its UV coordinate.
\end{enumerate}
The latter, instead, is used to improve texture quality as the SDS loss usually leads to artifacts that can be smoothed out. To do so, they render a blurry image $I^{p}_{\text{coarse}}$ from an arbitrary camera view $p$, and then they run a multi-step denoising process $f_{\Phi}(\cdot)$ using a Stable Diffusion 2D prior to get a refined image:
\begin{equation}
    I^{p}_{\text{fine}} = f_{\phi}\left(I^{p}_{\text{coarse}} + \epsilon(t_{\text{start}});t_{\text{start}}, \hat{c}_{T}\right)
\end{equation}
where $\epsilon(t_{\text{start}})$ is a random noise at timestamp $t_{\text{start}}$ and $\hat{c}_{T}$ is the (generative) conditioning. 
Since GSEdit uses Stable Diffusion as a guidance model during texture refinement, a small change in the prompt is required to give the model the correct conditioning. For instance, if our original GSEdit query was an elephant together with the instruction ``make it look like a mammoth'',  the texture refiner stage will require the edited object and a prompt that will work for a generative model, such as ``a mammoth''.
The refined image is used to optimize the texture through a pixel-wise MSE loss:
\begin{equation}
    \mathcal{L}_{\text{MSE}} = \left\lVert I^{p}_{\text{fine}} - I^{p}_{\text{coarse}} \right\rVert^{2}
\end{equation}

\begin{figure}[H]
    \centering
    \includegraphics[width=.32\linewidth]{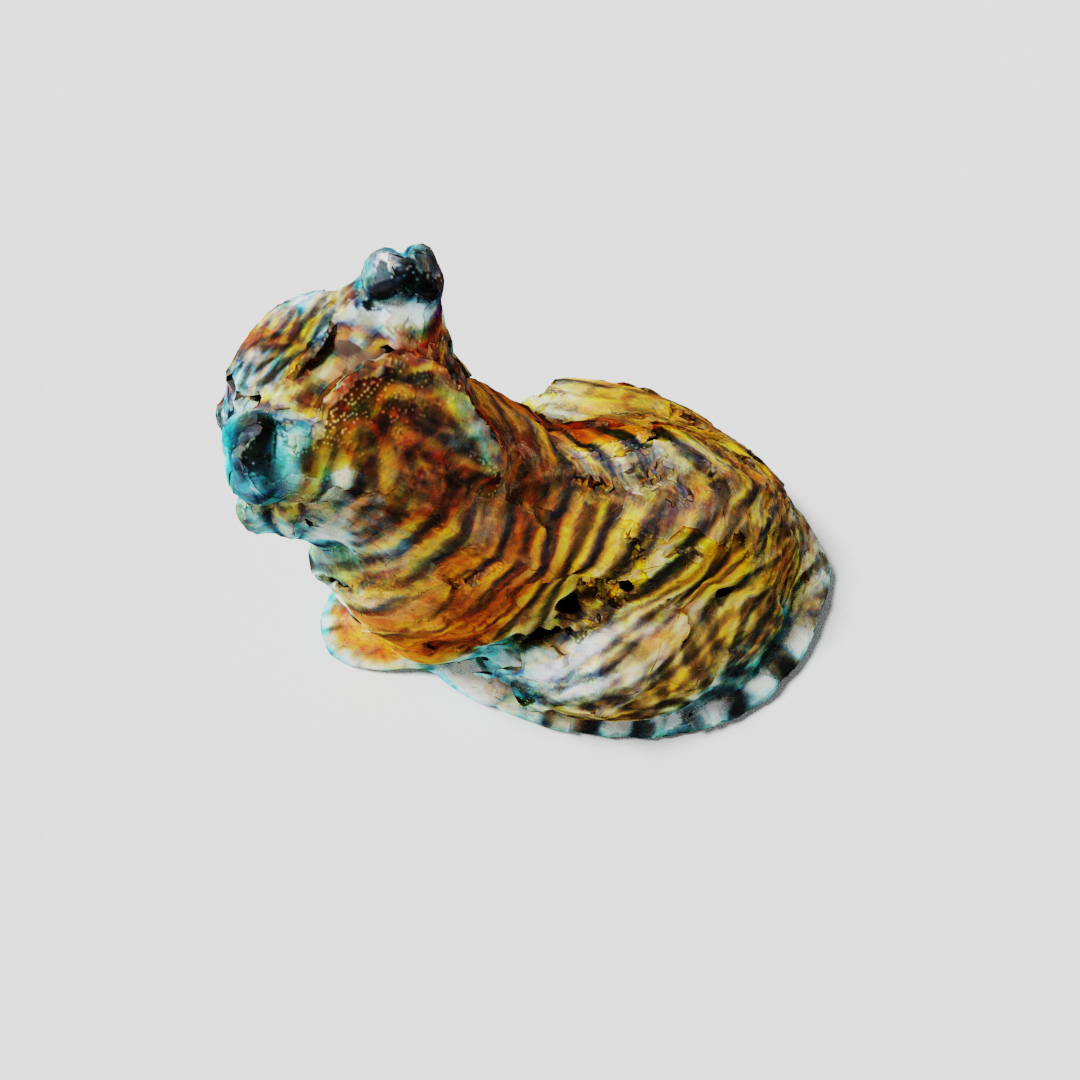}
    \includegraphics[width=.32\linewidth]{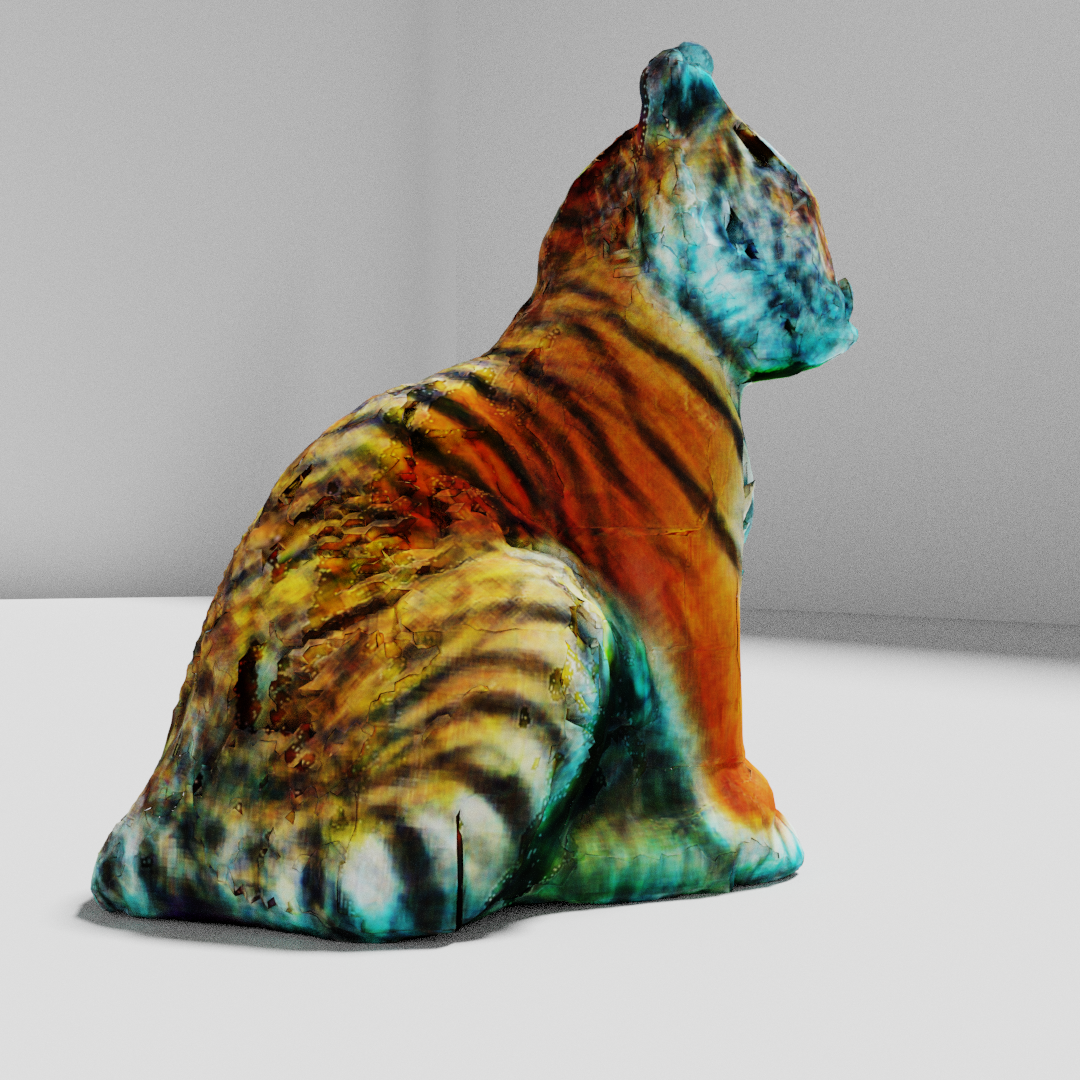}
    \includegraphics[width=.32\linewidth]{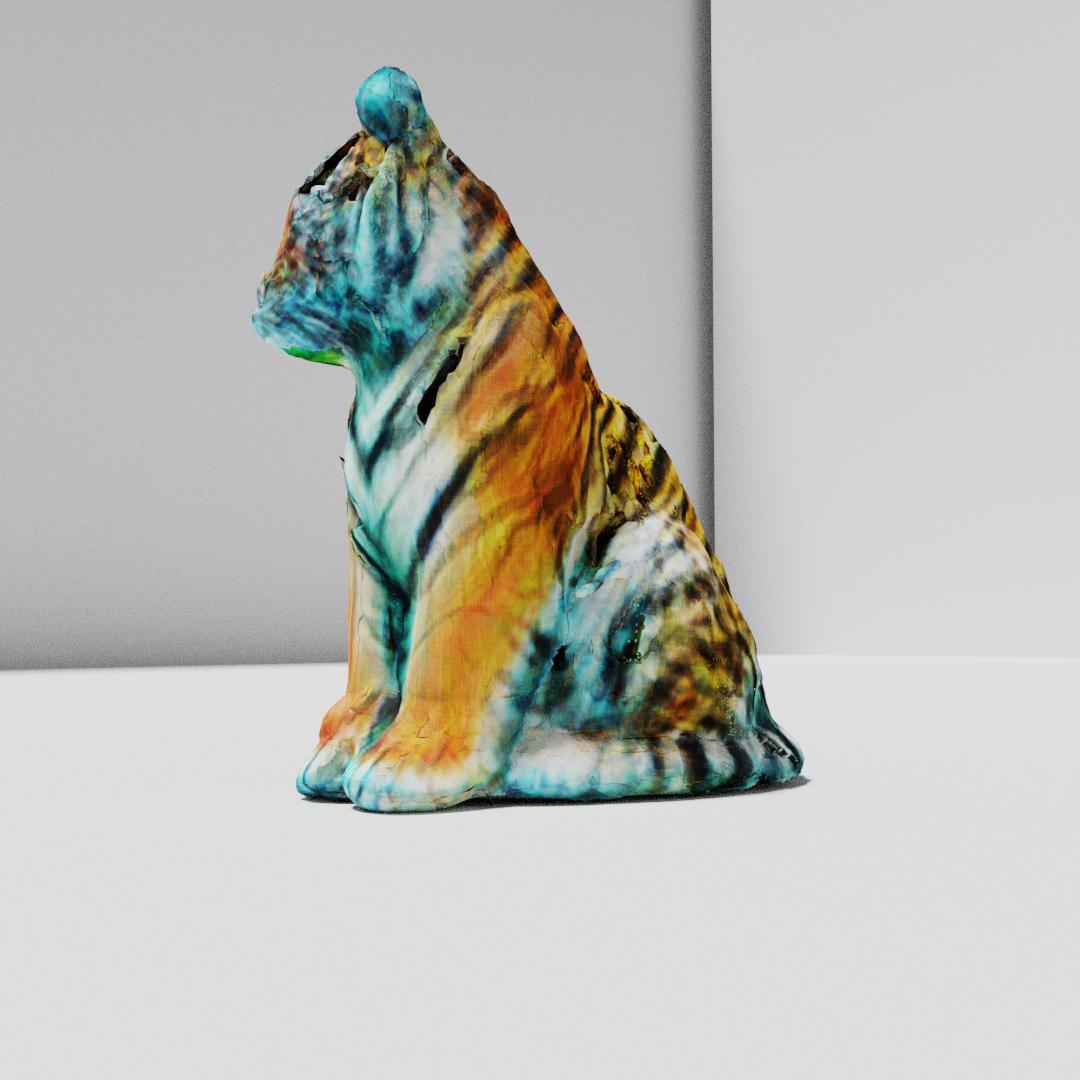}
    
    \includegraphics[width=.32\linewidth]{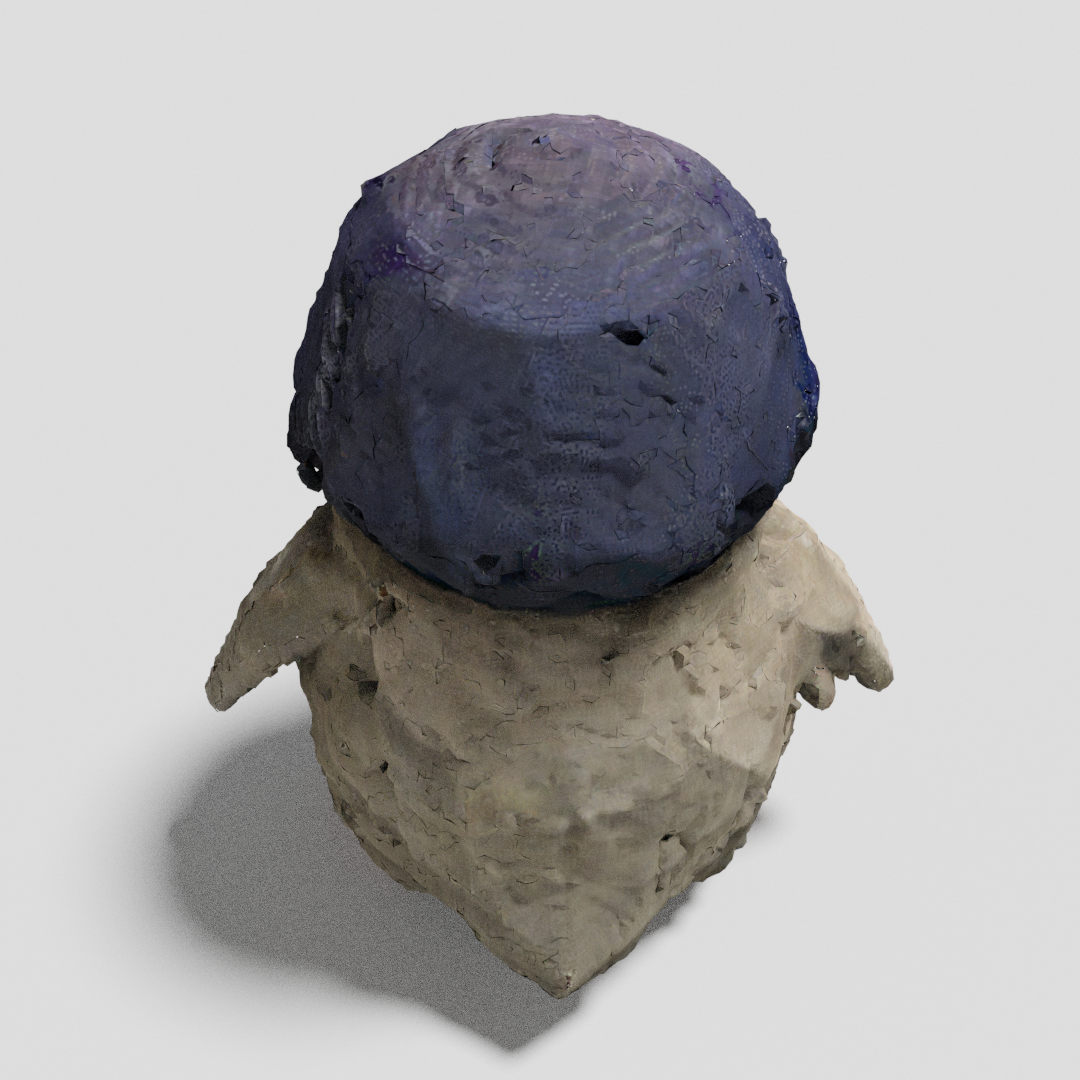}
    \includegraphics[width=.32\linewidth]{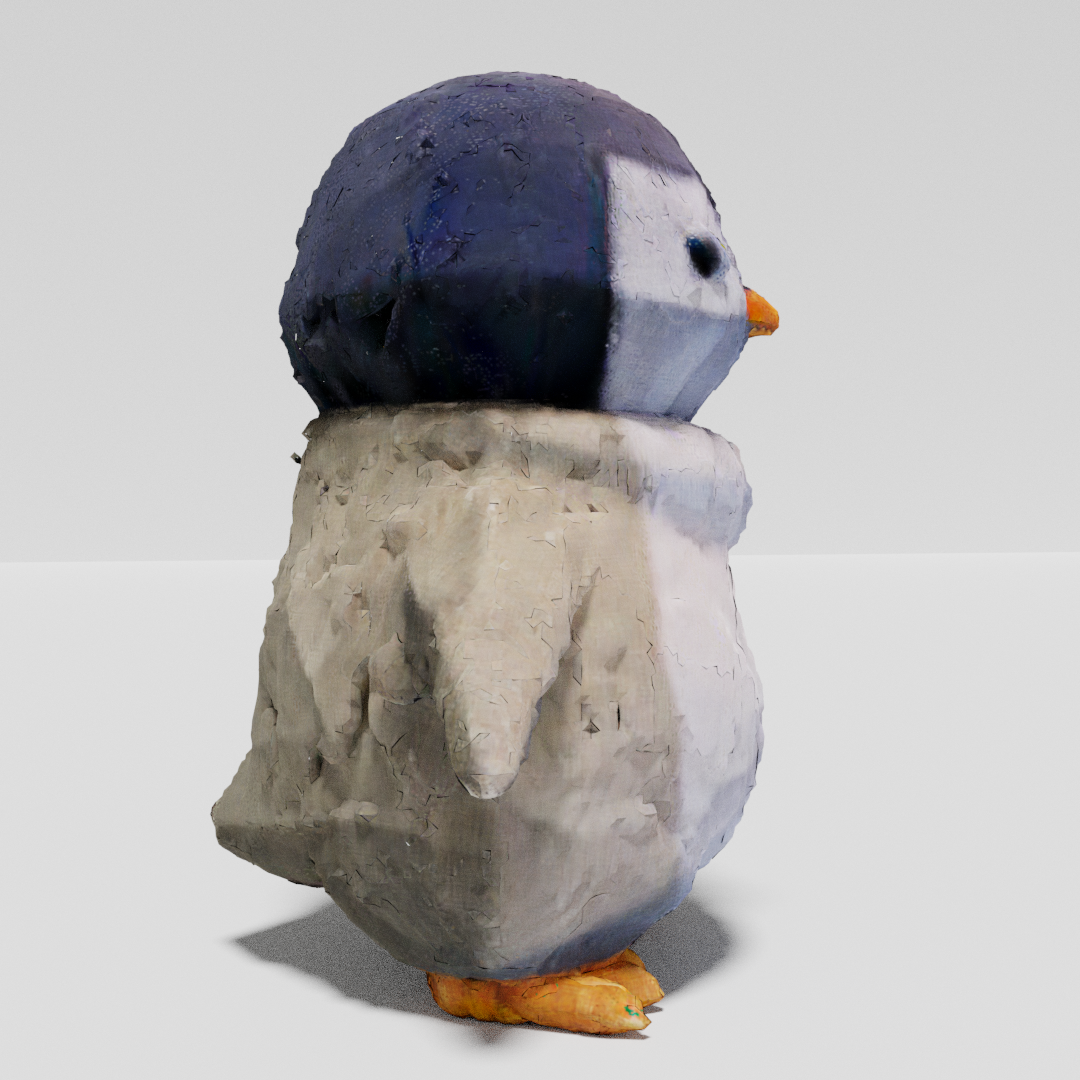}
    \includegraphics[width=.32\linewidth]{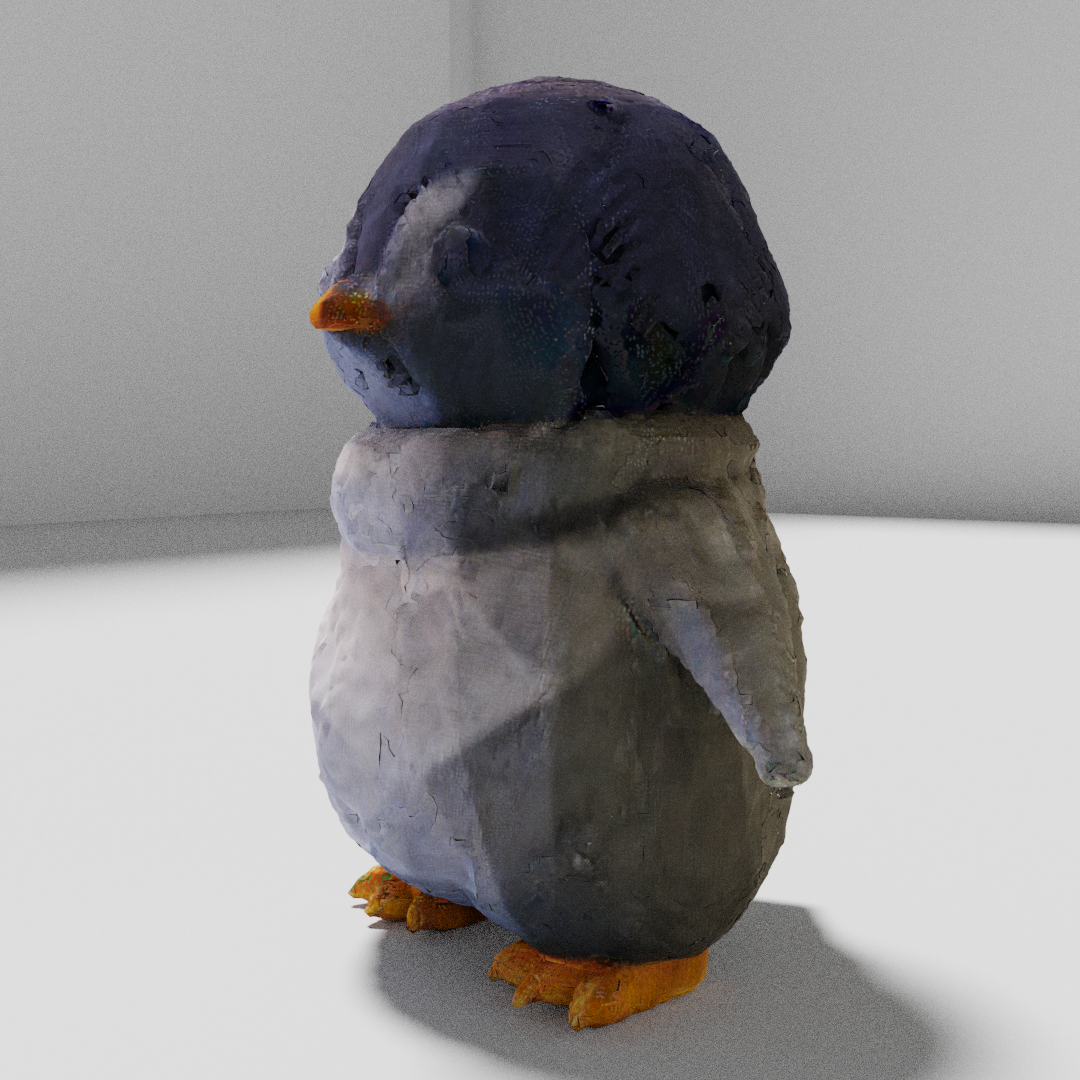}
    
    \includegraphics[width=.32\linewidth]{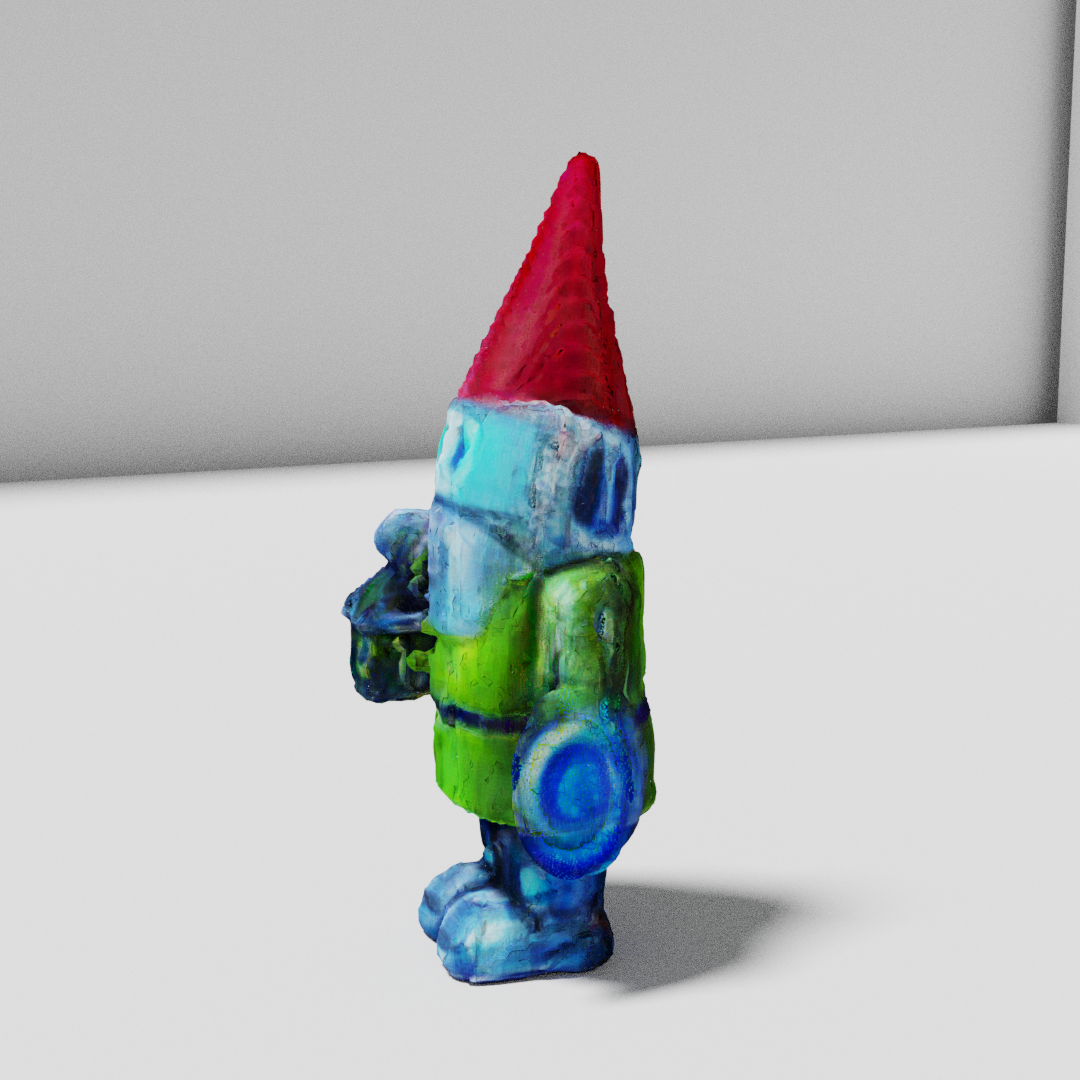}
    \includegraphics[width=.32\linewidth]{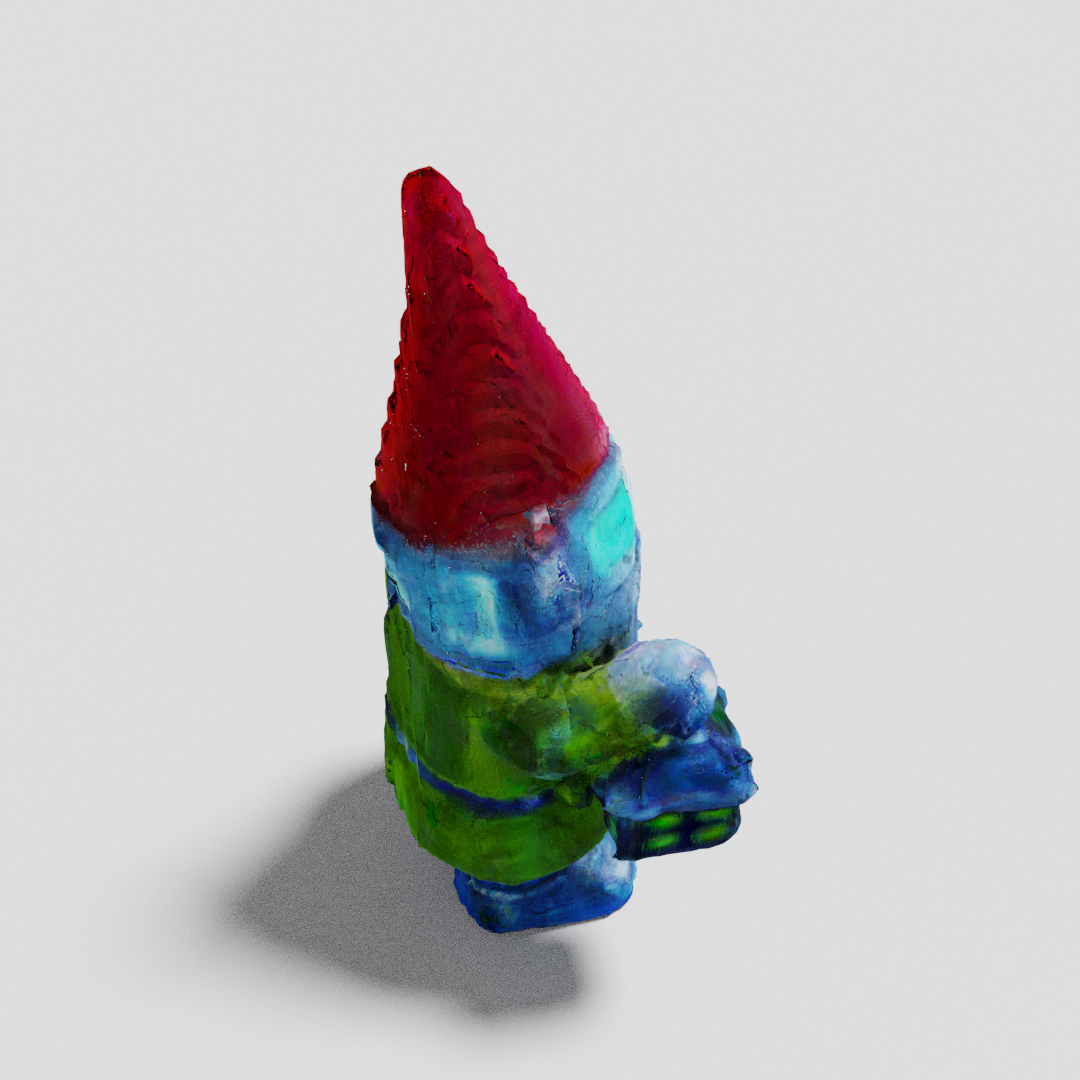}
    \includegraphics[width=.32\linewidth]{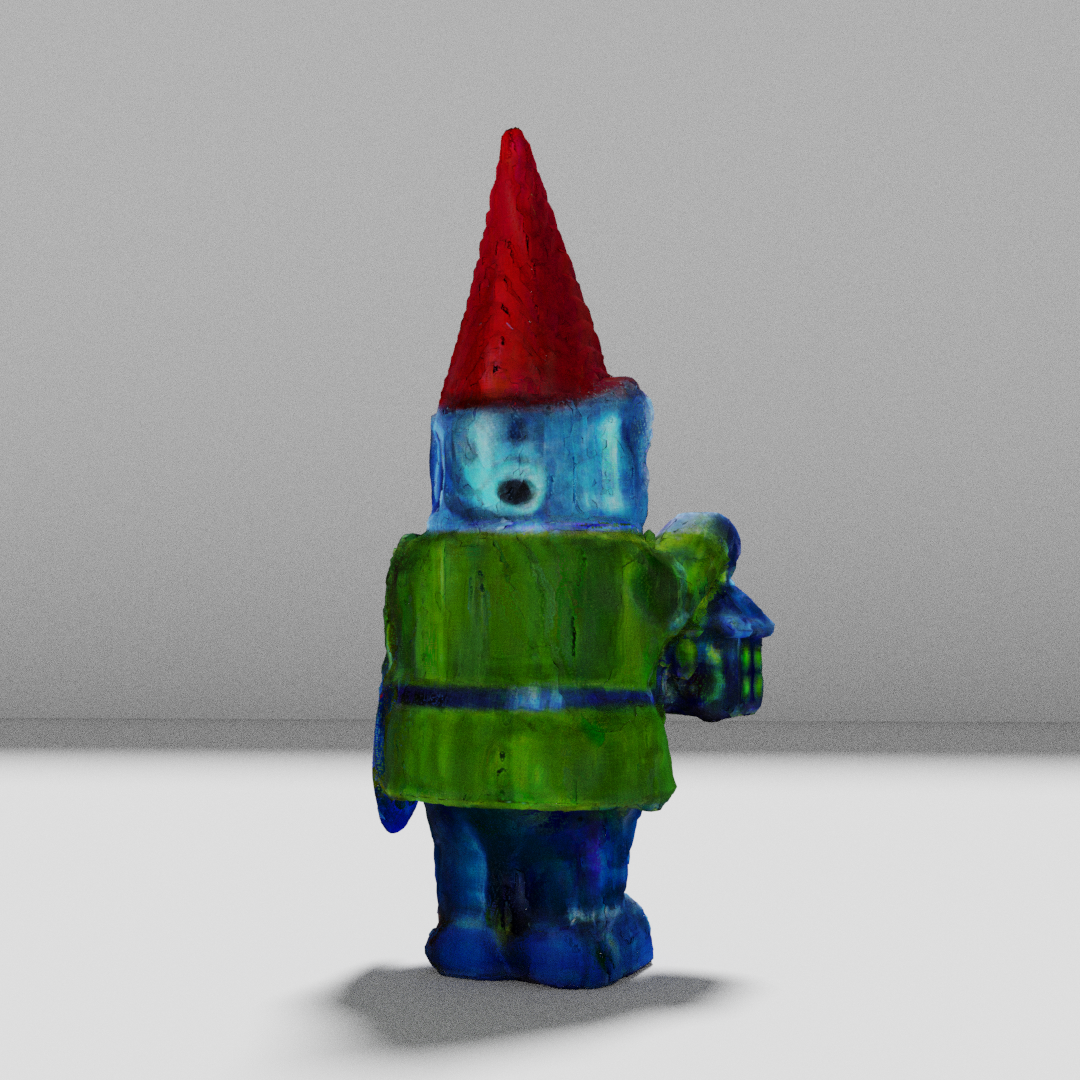}
    \caption{
    Multi-view renders of three meshes edited with GSEdit. 
    The results show good consistency of the edited shapes with respect to the view direction.}
    \label{fig:multi-view-render}
\end{figure}

\begin{figure*}[h]
    \centering
    \begin{tabular}{c c c c c c}
        Input shape & Prompt & IN2N & Vox-E & GaussianEditor &\textbf{GSEdit (ours)} \\
        \includegraphics[width=.145\textwidth]{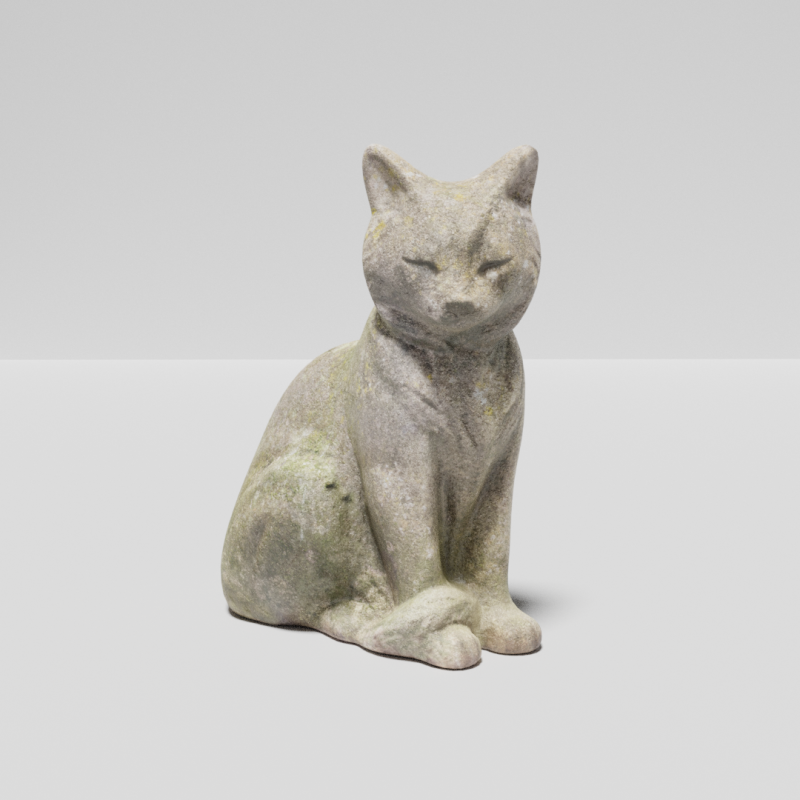} &
        \includegraphics[width=.145\textwidth]{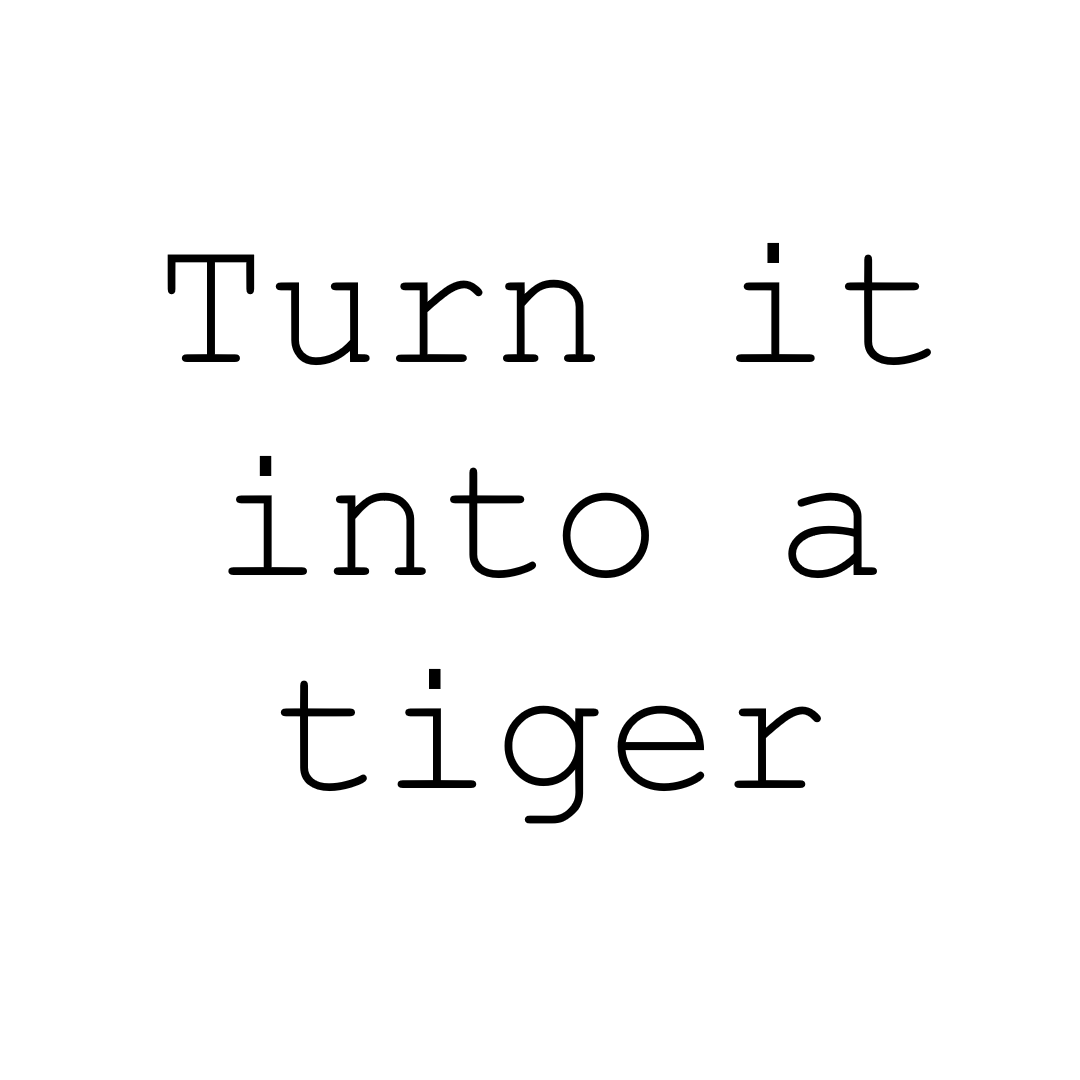} & 
        \begin{picture}(.145\textwidth, .145\textwidth)
            \put(0,0){\includegraphics[width=.145\textwidth]{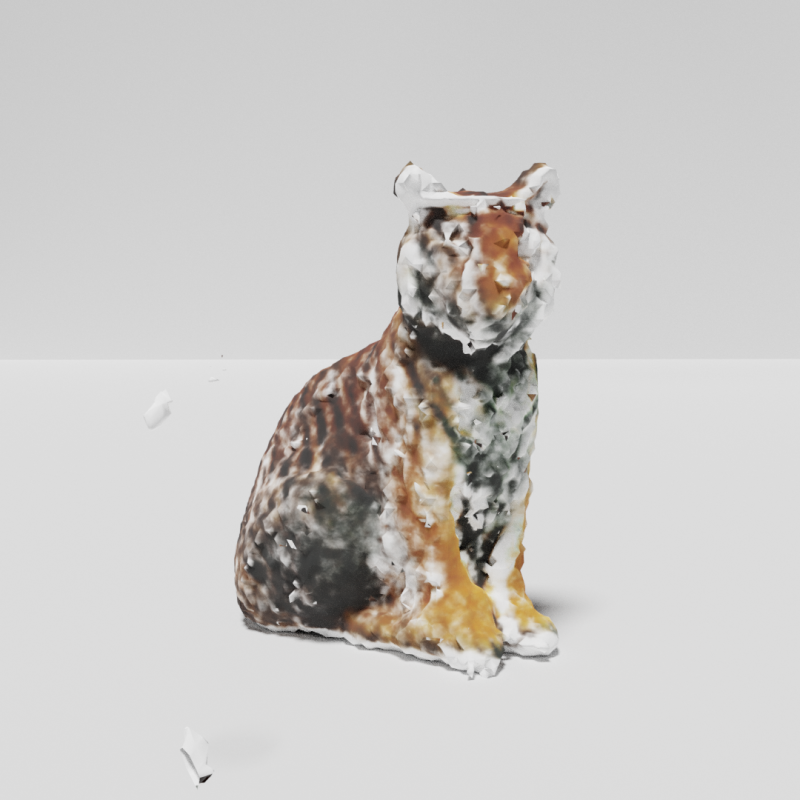}}
            \put(5, 5){$\sim $23 min}
        \end{picture}
        & 
        \begin{picture}(.145\textwidth, .145\textwidth)
            \put(0,0){\includegraphics[width=.145\textwidth]{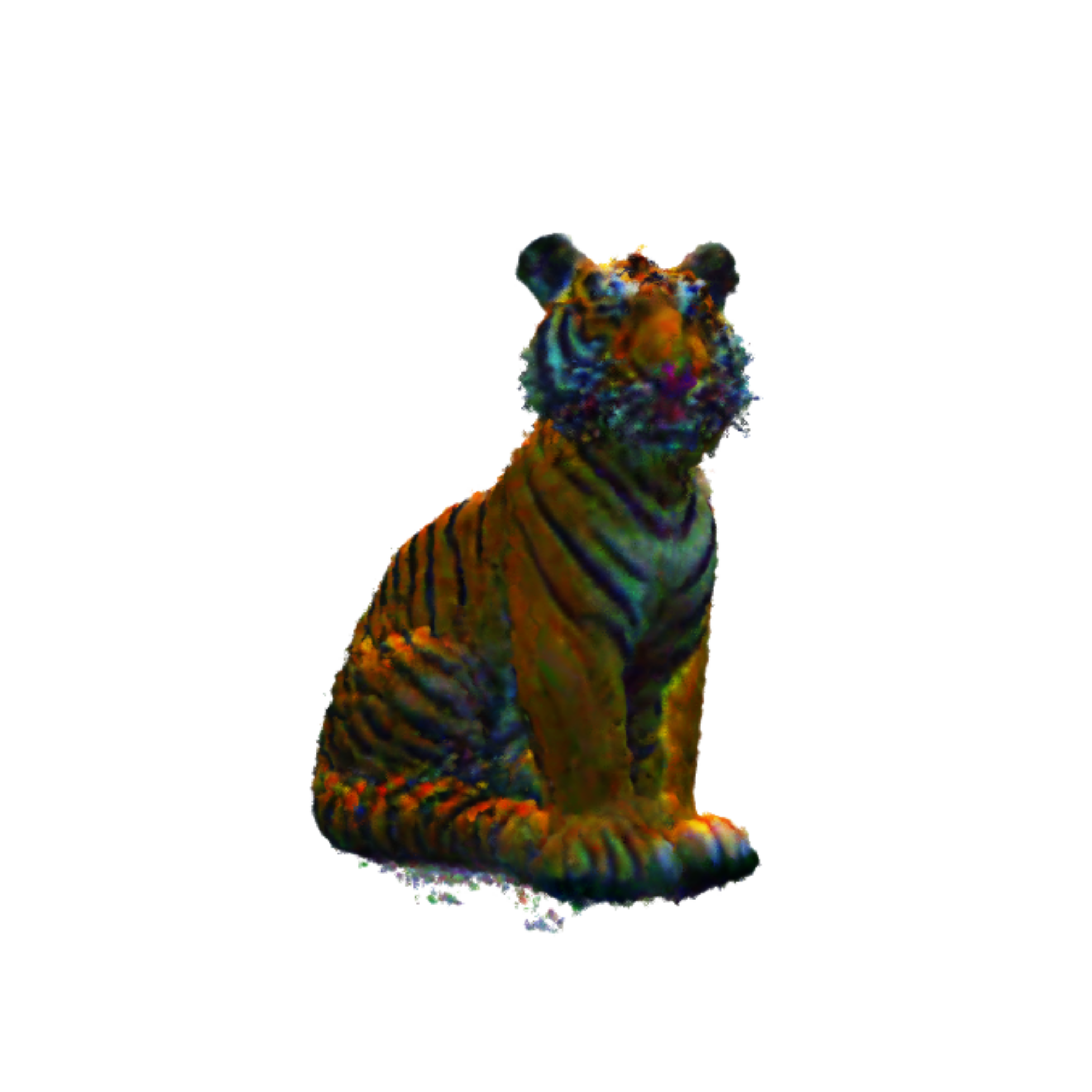}}
            \put(5, 5){$\sim $50 min}
        \end{picture}
        &
        \begin{picture}(.145\textwidth, .145\textwidth)
            \put(0,0){\includegraphics[width=.145\textwidth]{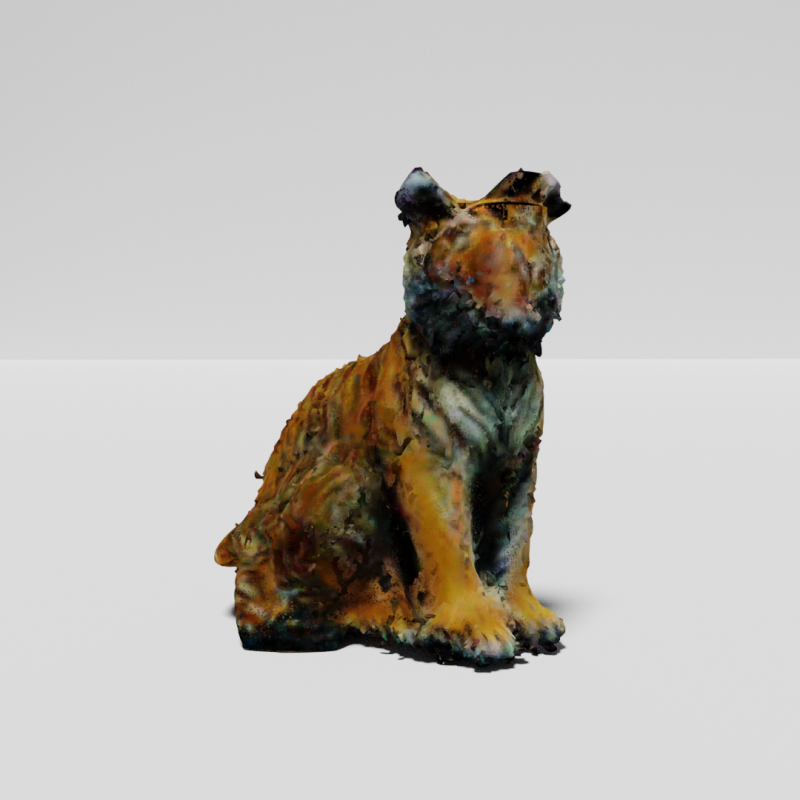}}
            \put(5, 5){$\sim$6.5 min}
        \end{picture}
        &
        \begin{picture}(.145\textwidth, .145\textwidth)
            \put(0,0){\includegraphics[width=.145\textwidth]{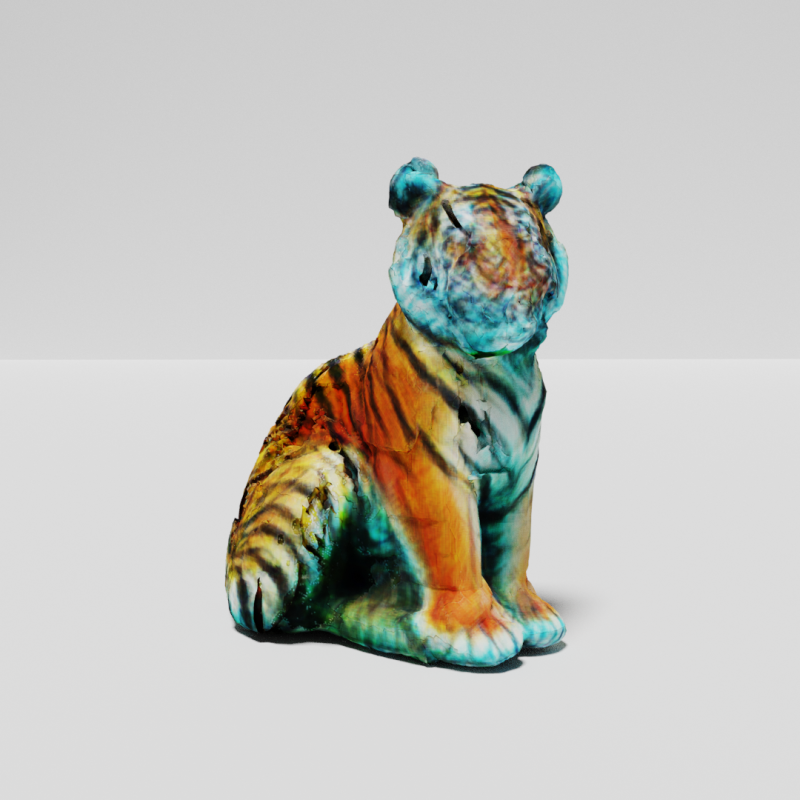}}
            \put(5, 5){$\sim$4 min}
        \end{picture}
        \\        
        \includegraphics[width=.145\textwidth]{imgs/data/data_penguin.png} &
        \includegraphics[width=.145\textwidth]{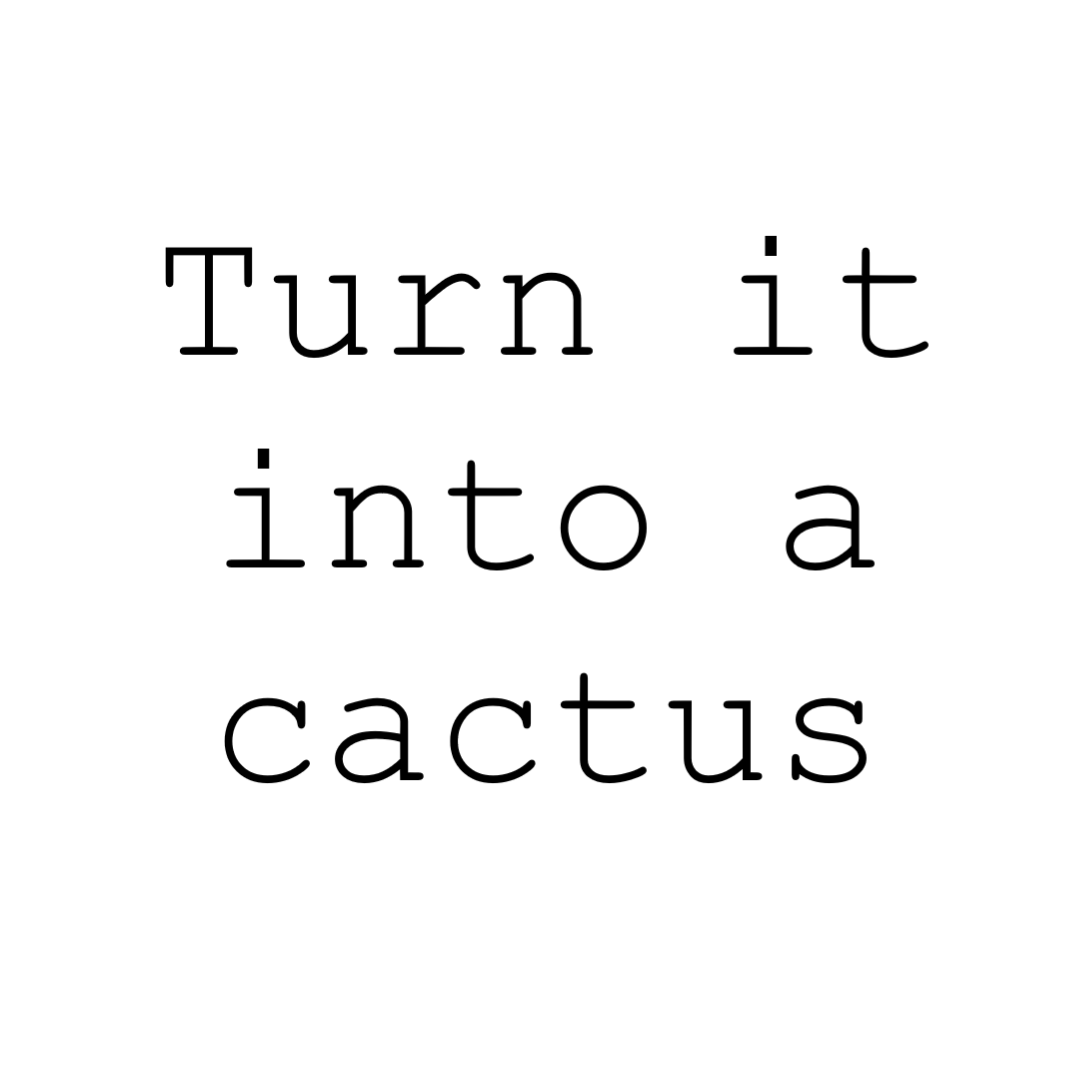} &
        \begin{picture}(.145\textwidth, .145\textwidth)
            \put(0,0){\includegraphics[width=.145\textwidth]{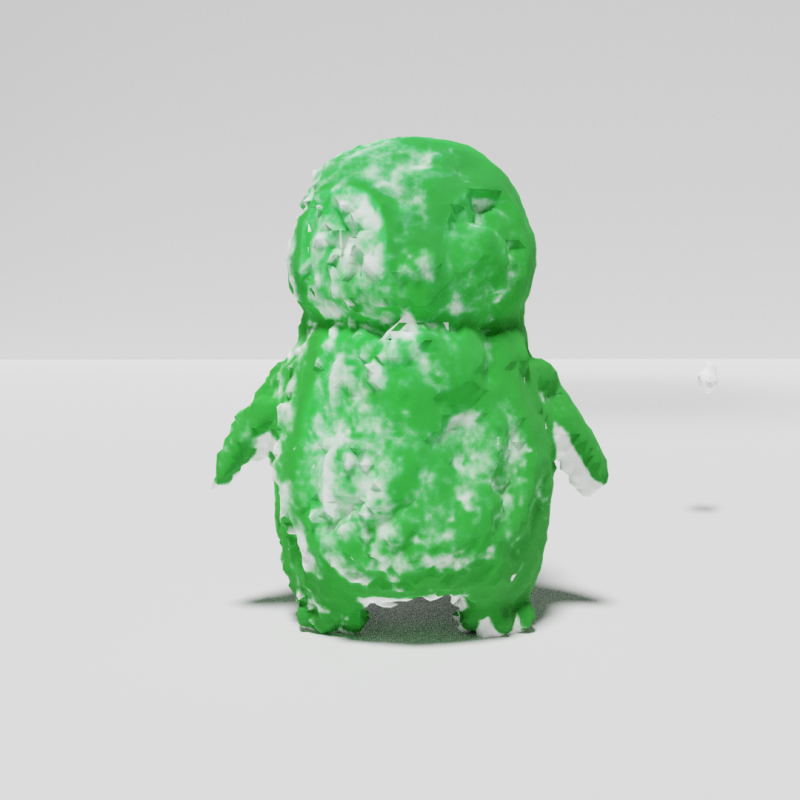}}
            \put(5, 5){$\sim $20 min}
        \end{picture}
        & 
        \begin{picture}(.145\textwidth, .145\textwidth)
            \put(0,0){\includegraphics[width=.145\textwidth]{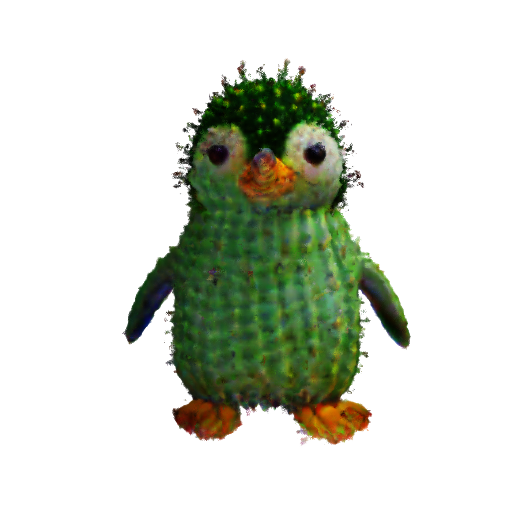}}
            \put(5, 5){$\sim $50 min}
        \end{picture}
        & 
        \begin{picture}(.145\textwidth, .145\textwidth)
            \put(0,0){\includegraphics[width=.145\textwidth]{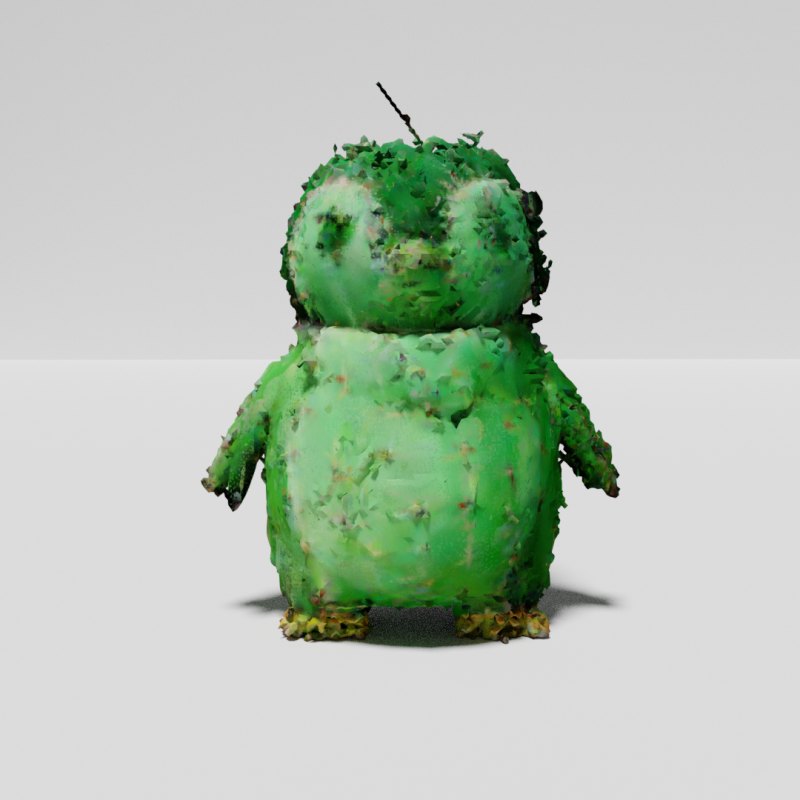}}
            \put(5, 5){$\sim $6.5 min}
        \end{picture}
        & 
        \begin{picture}(.145\textwidth, .145\textwidth)
            \put(0,0){\includegraphics[width=.145\textwidth]{imgs/ours/edit_penguin_cactus.png}}
            \put(5, 5){$\sim $4 min}
        \end{picture}
        \\        
        \includegraphics[width=.145\textwidth]{imgs/data/data_dog.png} &
        \includegraphics[width=.145\textwidth]{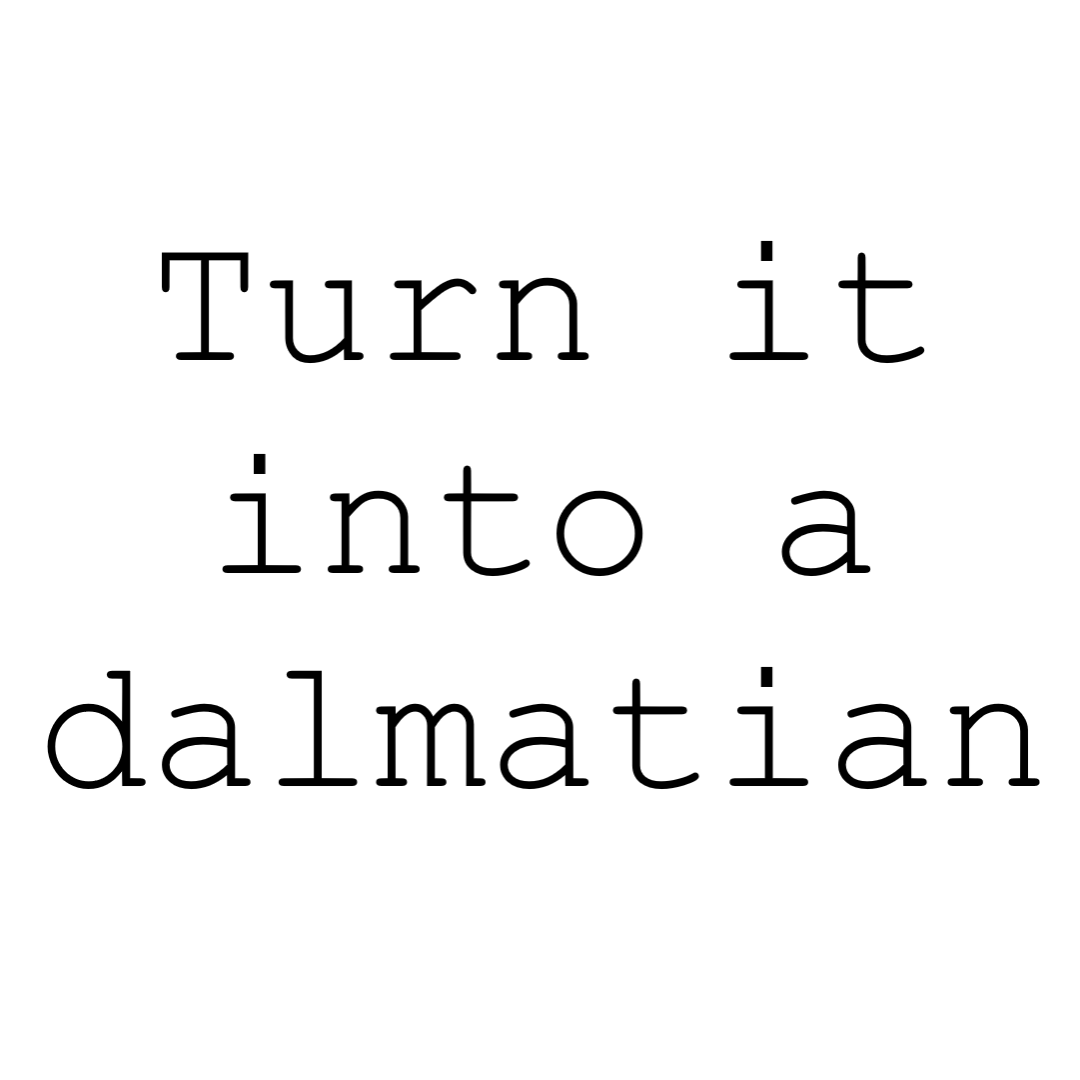} & 
        \begin{picture}(.145\textwidth, .145\textwidth)
            \put(0,0){\includegraphics[width=.145\textwidth]{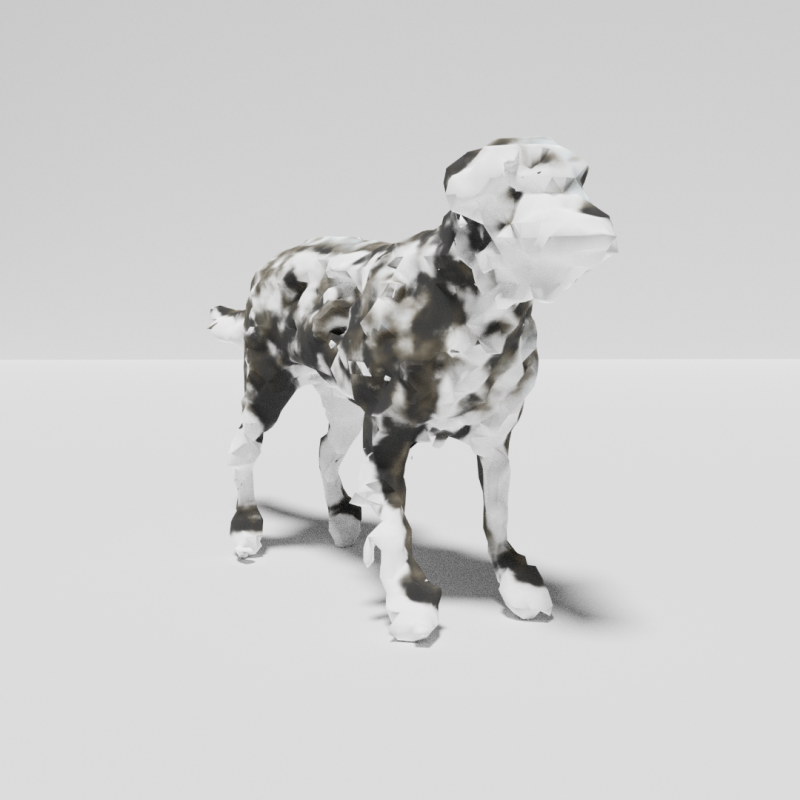}}
            \put(5, 5){$\sim $13 min}
        \end{picture}
        & 
        \begin{picture}(.145\textwidth, .145\textwidth)
            \put(0,0){\includegraphics[width=.145\textwidth]{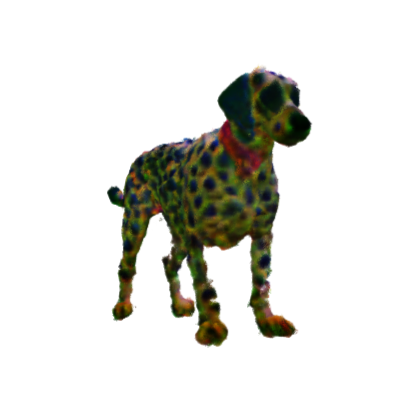}}
            \put(5, 5){$\sim $50 min}
        \end{picture}
        &
        \begin{picture}(.145\textwidth, .145\textwidth)
            \put(0,0){\includegraphics[width=.145\textwidth]{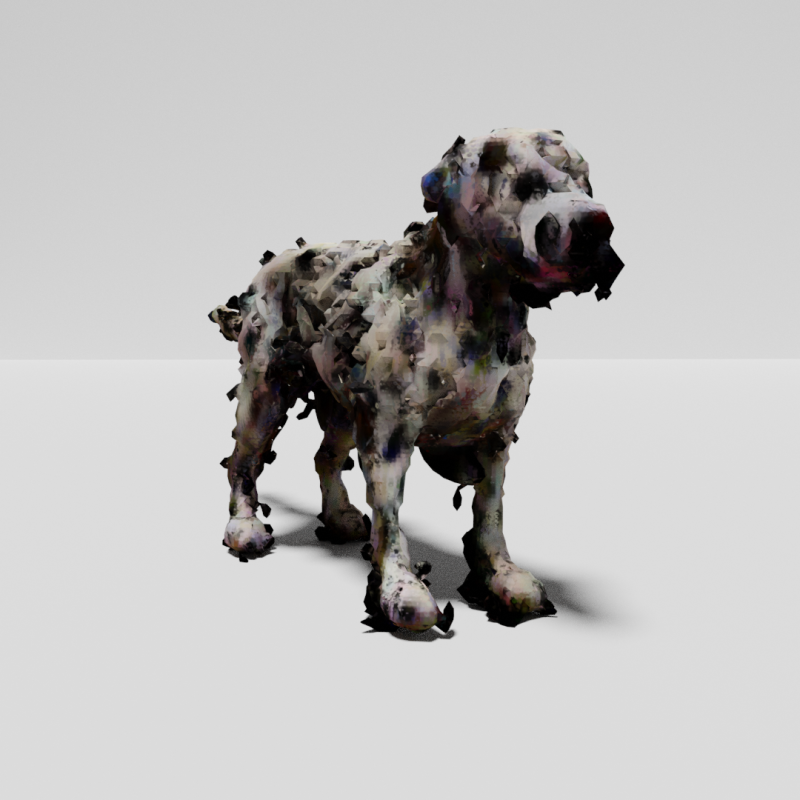}}
            \put(5, 5){$\sim $6.5 min}
        \end{picture}
        &
        \begin{picture}(.145\textwidth, .145\textwidth)
            \put(0,0){\includegraphics[width=.145\textwidth]{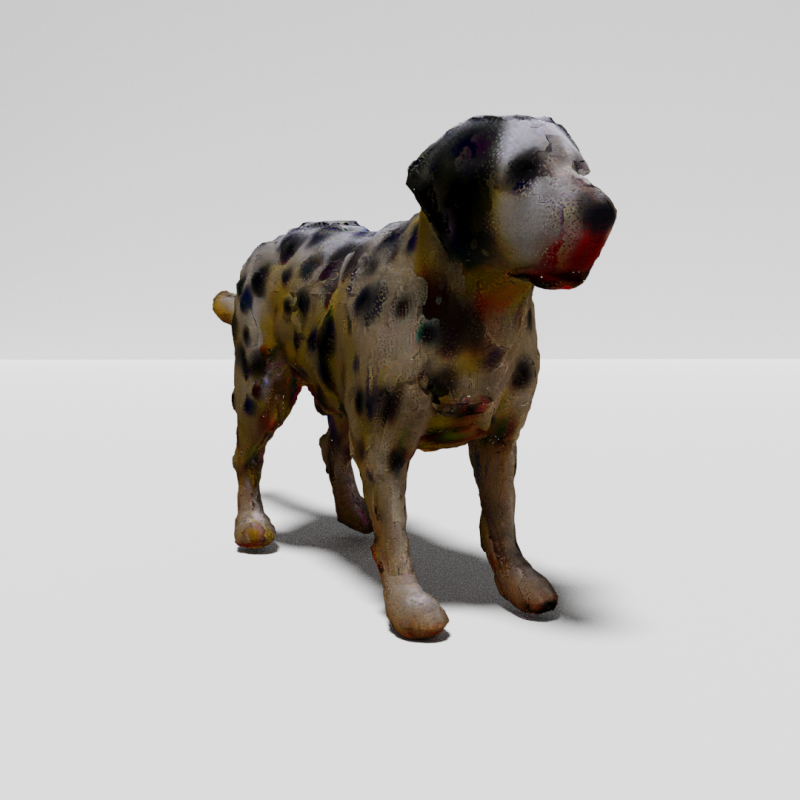}}
            \put(5, 5){$\sim $4 min}
        \end{picture}
        \\        
            \includegraphics[width=.145\textwidth]{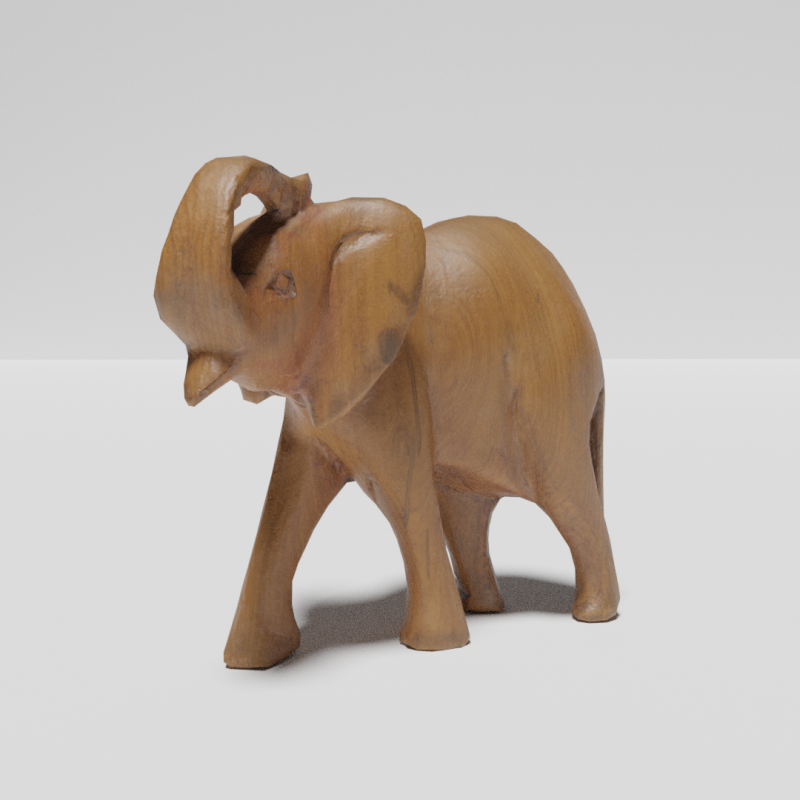} &
        \includegraphics[width=.145\textwidth]{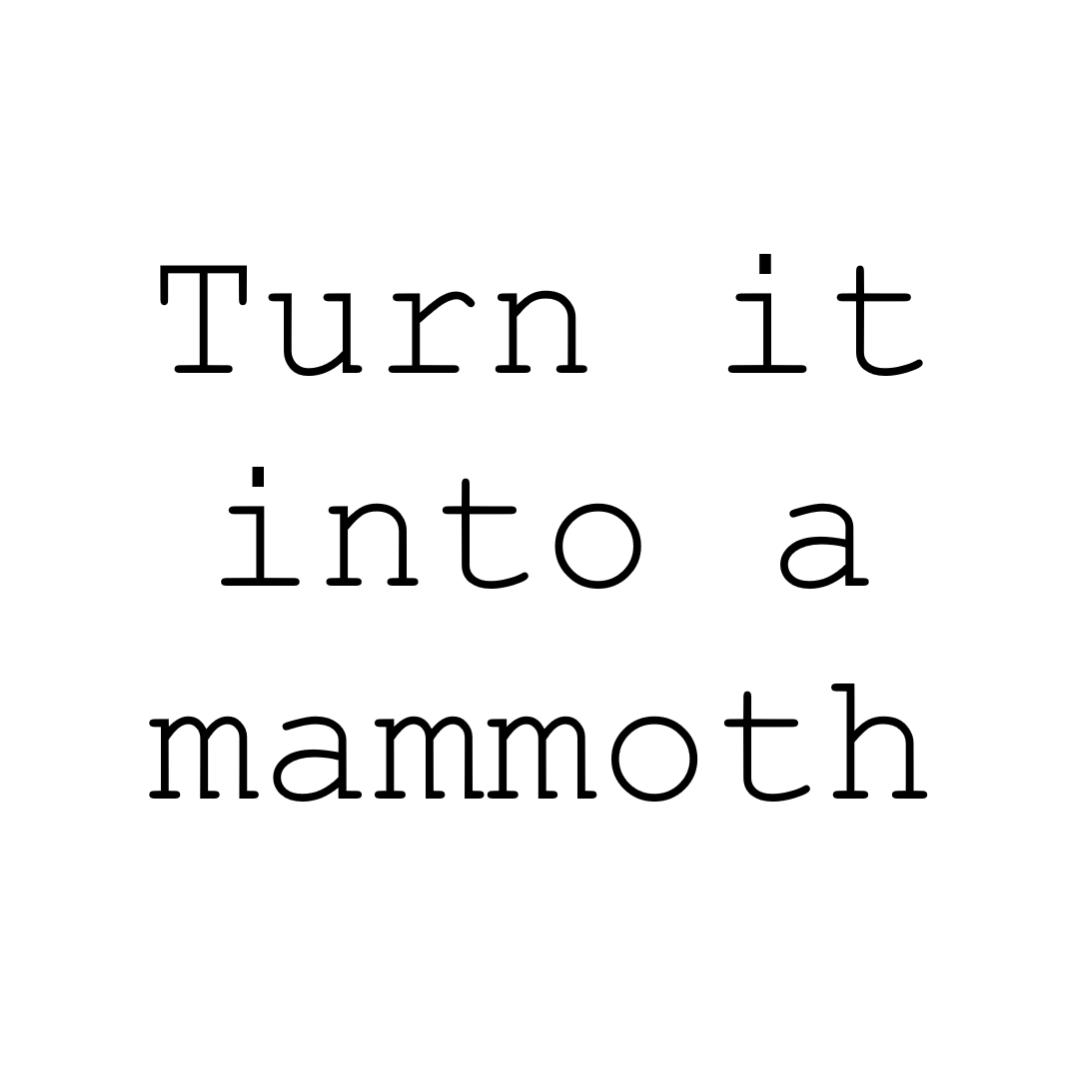} & 
        \begin{picture}(.145\textwidth, .145\textwidth)
            \put(0,0){\includegraphics[width=.145\textwidth]{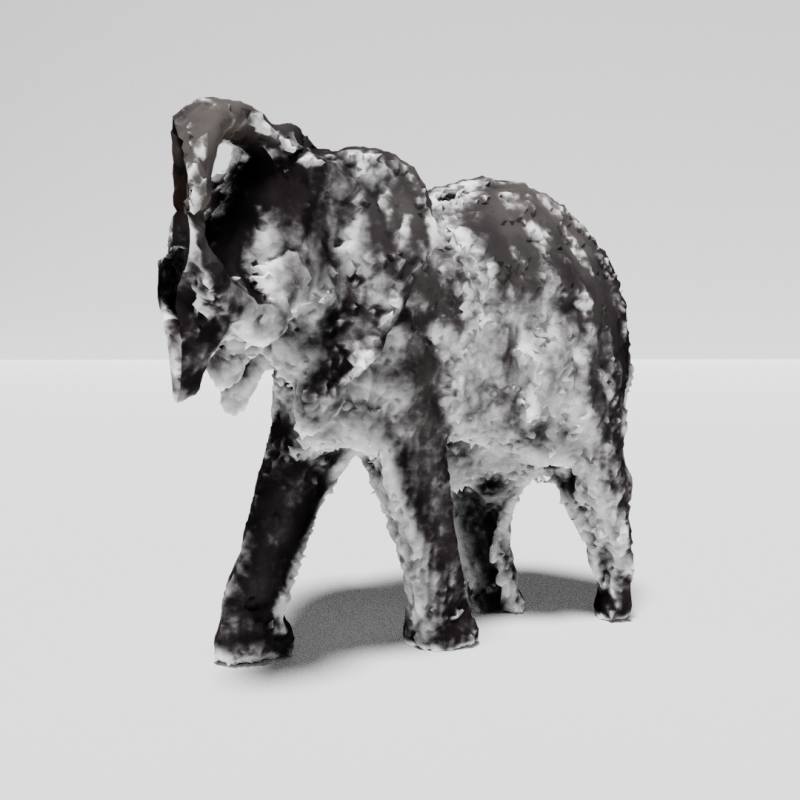}}
            \put(5, 5){$\sim $41 min}
        \end{picture}
        & 
        \begin{picture}(.145\textwidth, .145\textwidth)
            \put(0,0){\includegraphics[width=.145\textwidth]{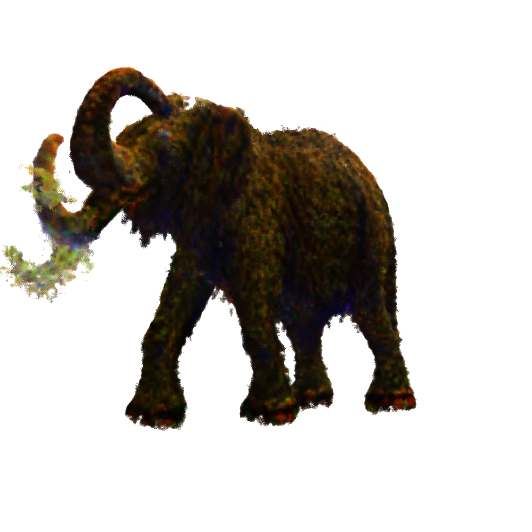}}
            \put(5, 5){$\sim $50 min}
        \end{picture}
        &
        \begin{picture}(.145\textwidth, .145\textwidth)
            \put(0,0){\includegraphics[width=.145\textwidth]{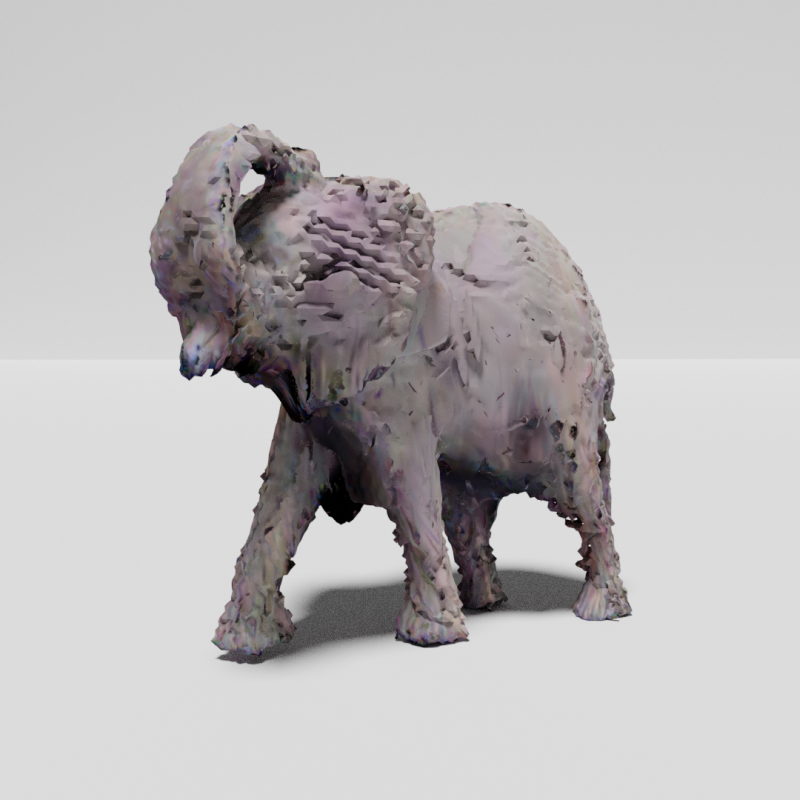}}
            \put(5, 5){$\sim $6.5 min}
        \end{picture}
        &
        \begin{picture}(.145\textwidth, .145\textwidth)
            \put(0,0){\includegraphics[width=.145\textwidth]{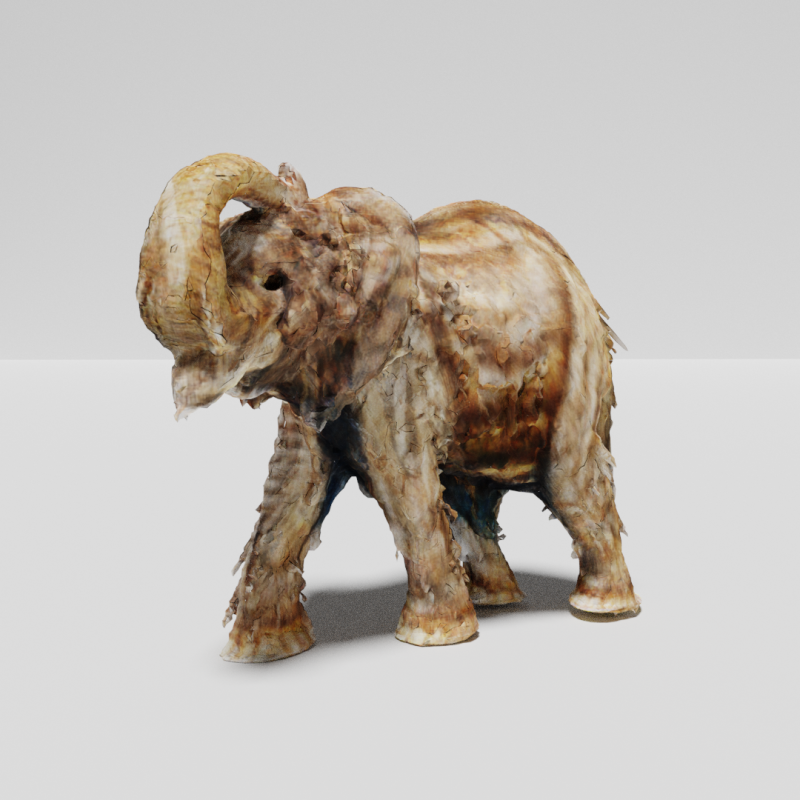}}
            \put(5, 5){$\sim $4 min}
        \end{picture}
        \\        
    \end{tabular}
    \caption{Qualitative comparison of results for object editing. The outputs of Instruct-NeRF2NeRF~\citet{instructnerf2023} show its inability to focus detail on a single object, hinting at the method only being suitable for entire scenes. Despite the additional contributions, the same weakness is partially inherited by GaussianEditor~\cite{chen2023gaussianeditor} due to employing the same backbone architecture. The results of Vox-E~\cite{voxe} can be more detailed than ours in some settings, however at a much higher computational cost and with no obvious way of extracting the geometry as a 3D mesh.}
    \label{fig:comparison-in2n}
\end{figure*}

\subsection{Implementation Details}
In our experimental setup, we initiated the reconstruction process with 10,000 Gaussians uniformly distributed within a cubic domain and positioned 20 cameras around the center. The locations of the cameras are expressed as spherical coordinates during setup and are distributed on two different elevations: 0 and 30 degrees, while the azimuth is a linear space between 0 and 360 degrees (last step excluded). Hence we have 10 cameras per elevation.  During the initial 3000 steps, the Gaussian model reconstructs the original object applying a densification step every 50 iterations.
For the editing, the hyper-parameters that control the editing are several and differ based on the object and the edit prompt. 
During the 500 iterations dedicated to editing, we vary editing intensity based on the specific prompts, adjusting the noise level parameters $[t_{min}, t_{max}]$ where we set $t_{min}=0.02$ and $t_{max}$ up to 0.98 for extensive modification cases. The rationale applied to the noise parameters similarly extends to the conditioning guidance scales for text, $s_T$, and image, $s_I$. These scales serve as control mechanisms to ensure the edited object retains the original shape's integrity. We set these parameters at $s_T = 100$ and $s_I = 10$, respectively, mirroring the configuration in DreamGaussian, with potential hand-tune adjustments for balancing original shape retention and desired edit strength.

\section{Results}

Due to the lack of an established benchmark dataset for the task at hand, we gathered a small set of textured 3D meshes from online repositories (all under CC-0/CC-BY copyright license) in order to evaluate our method against previous work. The artists will be eventually acknowledged in the final version of the paper. Evaluating GSEdit on heterogeneous artist-made 3D data has the additional benefit of showing the applicability of our method to real-world scenarios. All our experiments were run on an NVIDIA RTX 3090 24 GB VRAM, with an intel core i7-12700f and 32 GB DDR5 5600MT/s RAM.

\subsection{Qualitative Results}
Our qualitative results are shown in \Cref{fig:teaser,fig:collection,fig:comparison-in2n,fig:multi-view-render}. GSEdit shows remarkable capabilities in consistently editing Gaussian splatting object so that the final result can be extracted as a mesh.
It is possible to modify the shape of objects as well as the color and yield good results once the correct parameters are set. However, the editing power is heavily bound to the capabilities of IP2P (as we discuss later in \Cref{sec:conclusions}), so finding good parameters is fundamental to achieving the desired mesh as output. GSEdit is able to modify the shape, the color, and the overall style of the input meshes according to the input prompt. Our experiments include changes in object identity with preservation of original features, and in visual appearance, such as material or artistic style.

\subsection{Quantitative Results}
Text-guided editing is mostly a subjective task, in the sense that two distinct ``evaluators'' could expect different outcomes out of the same prompt: even the idea of a benchmark set is ill-defined in this setting. As a consequence, it is nontrivial to identify some metric that objectively expresses how accurate the modifications for given inputs are. Following previous work~\cite{instructnerf2023,qian2024magic,liu2023one}, we analyze the edited object in the CLIP~\cite{clip} space to compare it with the prompt. In particular, the two following CLIP-based metrics can be used.

\paragraph{CLIP directional similarity} This metric compares a render of the original 3D object $x^{(i)}$ from some viewpoint $i$ the and the render of the edited object $\hat{x}^{(i)}$ from the same viewpoint and two prompts, one describing the original object $T^{(i)}$ (\eg~\textit{``A cartoonish penguin''}) and describing the edited object $\hat{T}^{(i)}$(\eg~\textit{``A cartoonish penguin made of gold''}). Then, we embed the image pair and the text pair in the CLIP space, and we compute the cosine similarity between the offsets $\left(x_{\text{CLIP}}^{(i)} - \hat{x}_{\text{CLIP}}^{(i)}\right)$ and $\left(T_{\text{CLIP}}^{(i)} - \hat{T}_{\text{CLIP}}^{(i)}\right)$, as:
\begin{equation}
    \text{CLIP}_{\textit{sim}} = \dfrac{\left(x_{\text{CLIP}}^{(i)} - \hat{x}_{\text{CLIP}}^{(i)}\right) \cdot \left({T_{\text{CLIP}}^{(i)} - \hat{T}_{\text{CLIP}}^{(i)}}\right)}{\left\lVert x_{\text{CLIP}}^{(i)} - \hat{x}_{\text{CLIP}}^{(i)} \right\rVert \cdot \left\lVert T_{\text{CLIP}}^{(i)} - \hat{T}_{\text{CLIP}}^{(i)} \right\rVert}
\end{equation}

\paragraph{CLIP directional consistency}
In addition to evaluating the directional similarity, assessing the consistency of the edited object across camera views is imperative.  
To quantify this, we focus on analyzing pairs of adjacent frames, capturing the original and edited 3D objects from consecutive viewpoints along a camera path, denoted by $x^{(i)}, x^{(i+1)}$ for the original, and $\hat{x}^{(i)}, \hat{x}^{(i+1)}$ for their edited counterparts. Embedding these frames within the CLIP space allows us to easily evaluate the alignment of the editing across them. As for directional similarity, we obtain the CLIP directional consistency metric proposed by~\citet{instructnerf2023} as the cosine similarity between the change in the CLIP embeddings from one frame to the next:
\begin{equation}
    \text{CLIP}_{\textit{cons}} = \dfrac{\left(\hat{x}_{\text{CLIP}}^{(i)} - x_{\text{CLIP}}^{(i)}\right) \cdot \left({\hat{x}_{\text{CLIP}}^{(i+1)} - x_{\text{CLIP}}^{(i+1)}}\right)}{\left\lVert \hat{x}_{\text{CLIP}}^{(i)} - x_{\text{CLIP}}^{(i)} \right\rVert \cdot \left\lVert {\hat{x}_{\text{CLIP}}^{(i+1)} - x_{\text{CLIP}}^{(i+1)}} \right\rVert}
\end{equation}

Additionally, we report $\text{CLIP}_{\textit{text}}$, the CLIP text similarity, which evaluates to the cosine similarity between the CLIP embeddings of the generative prompt (\ie, ``a mammoth'' rather than ``turn it into a mammoth'') and the CLIP embeddings of the edited object render. Results for all metrics and execution time are reported in \Cref{tab:comparison_in2n}. 


\begin{table}[h]
\centering
\caption{Evaluation of quality and performance metrics for our method and multiple baselines: IN2N and GaussianEditor are scene editing methods, while Vox-E is focused on 3D objects, like our method. GSEdit presents a clear superiority in terms of the $\text{CLIP}_{\textit{sim}}$ metric ($\sim2\times$ with respect to all baselines) and runtime (speedups of $6.25\times, 1.65\times, 13.40\times$ respectively). This highlights the efficiency and the coherence of the editing provided by our method.
For metrics where it does not exceed the baselines scores, it still provides competitive results, while the same does not hold for the other three methods.}
\label{tab:comparison_in2n}

\begin{tabular}{l || c | c | c | c }
     & {IN2N}   & {GaussianEditor}  & {Vox-E} & \textbf{GSEdit} \\
      \hline
     $\text{CLIP}_{\textit{sim}} (\uparrow) $   & 0.1382 & 0.1313 & 0.1196 & \textbf{0.2345}  \\
     $\text{CLIP}_{\textit{cons}} (\uparrow)$  & 0.8226 & \textbf{0.8436} & 0.6500 & 0.8147  \\
     $\text{CLIP}_{\textit{text}} (\uparrow)$  & 0.1986 & 0.2236 & \textbf{0.2868} & 0.2631 \\
     Time (s)   & 1493   & 394  & 3204  & \textbf{239}

\end{tabular}
\end{table}

\begin{figure}[h]
    \centering
    \includegraphics[width=.32\linewidth]{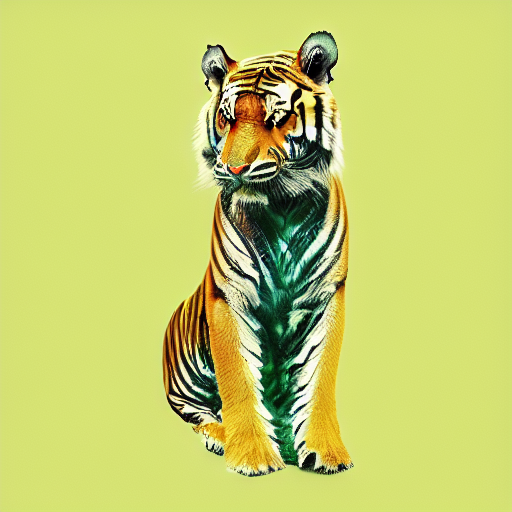} 
    \includegraphics[width=.32\linewidth]{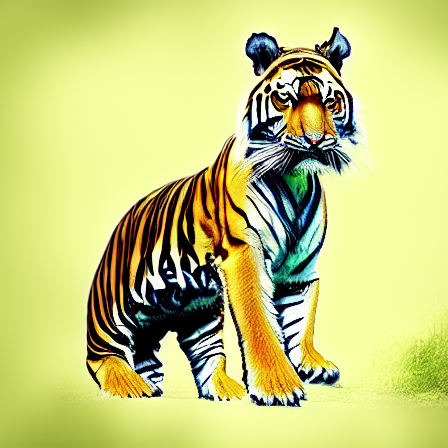}
    \includegraphics[width=.32\linewidth]{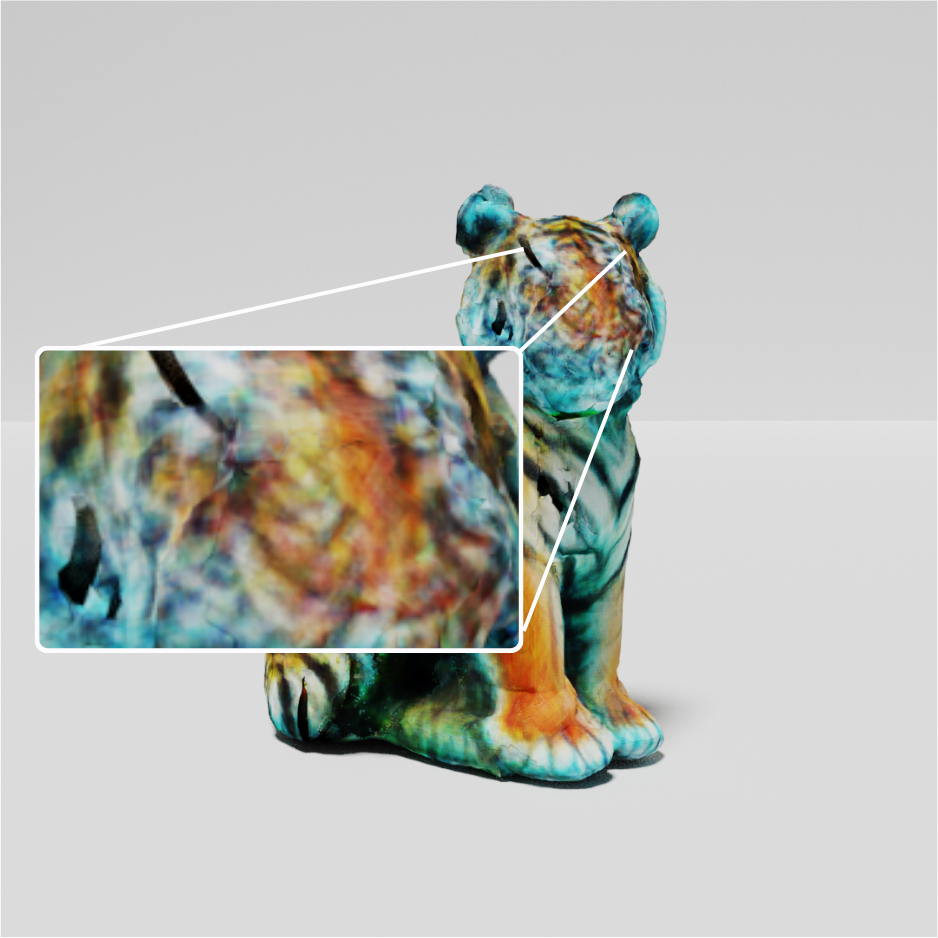}
    \caption{Illustration of Instruct-Pix2Pix limitations: Observed from two distinct angles, the tiger consistently appears to face the camera due to IP2P's processing. Such uniform orientation across various viewpoints leads to a blurred effect in the eye region upon reconstruction.}
    \label{fig:ip2p-limit}
\end{figure}

\section{Conclusions}\label{sec:conclusions}

We introduced GSEdit, a method for efficient text-guided editing of 3D objects based on image diffusion models and the Gaussian splatting scene representation. Our method is simple, building on top of previous research, and it allows to solve a complex problem much more efficiently than previous contributions, without sacrificing the quality of its output: in fact, our model yields qualitatively superior outputs to state of the art methods which, rather than focusing on 3D objects, tackle the problem for entire scenes, and it does so in a fraction of the time. While a baseline specifically for 3D object editing provides more competitive results to our method, the difference in editing time in this case increases dramatically, proving the usefulness of our contribution.
As we previously outlined, our contribution has significant applications in the fields of 3D vision and graphics, and it leaves room for improvement in future research: one particularly interesting avenue would be to find new ways to increase the view-wise consistency of the diffusion model's edits of the current scene views.

\paragraph{Limitations}
The methodology we employed is fundamentally dependent on the Instruct-Pix2Pix framework, which introduces certain constraints. As illustrated in \cref{fig:ip2p-limit}, a notable limitation is the perspective bias—regardless of the viewpoint, the tiger appears to be gazing directly at the camera, leading to an unnatural blurriness in the eyes within the final rendering. Additionally, the inherent capabilities of Instruct-Pix2Pix restrict substantial spatial transformations, such as altering poses or adding/removing objects, limiting the scope of possible edits. Moreover, the strategy of adjusting individual views can inadvertently result in artifacts, compromising the overall consistency and quality of the edited output.

\bibliographystyle{ACM-Reference-Format}
\bibliography{sample-base}

\end{document}